\def\year{2020}\relax
\documentclass[letterpaper]{article} %
\usepackage{aaai20}  %
\usepackage{times}  %
\usepackage{helvet} %
\usepackage{courier}  %
\usepackage[hyphens]{url}  %
\usepackage{graphicx} %
\def\UrlFont{\rm}  %
\usepackage{graphicx}  %
\nocopyright
\title{Deep Conservative Policy Iteration}

\author{\textbf{Nino Vieillard, Olivier Pietquin, Matthieu Geist}\\ %
\\
Google Research, Brain Team
}

\usepackage[utf8]{inputenc} %
\usepackage[T1]{fontenc}    %
\usepackage{amsthm}
\usepackage{url}            %
\usepackage{booktabs}       %
\usepackage{amsfonts}       %
\usepackage{nicefrac}       %
\usepackage{microtype}      %
\usepackage{mathtools}
\usepackage{lmodern}
\usepackage{amsmath}
\usepackage{algorithmic}
\usepackage{algorithm}
\usepackage{placeins}
\usepackage{bbold}

\DeclareMathOperator*{\argmax}{argmax}

\DeclareMathOperator{\kl}{KL}
\DeclareMathOperator{\FC}{FC}
\DeclareMathOperator{\Conv}{Conv}
\DeclareMathOperator*{\avg}{\hat{\mathbb{E}}}
\newcommand{\gr}{\mathcal{G}}
\newcommand{\p}[1]{#1^{\prime}}
\newcommand{\old}[1]{#1^{-}}
\newcommand{\adv}{A_{\pi}^{\bar\pi}}
\newcommand{\advmu}{A_{\pi, \mu}^{\bar\pi}}
\newcommand{\states}{\mathcal{S}}
\newcommand{\actions}{\mathcal{A}}

\mathtoolsset{showonlyrefs=true}

\newtheorem{theorem}{Theorem}

\newcommand{\citet}[1]{\citeauthor{#1} \shortcite{#1}}

\begin{document}
\maketitle

\begin{abstract}
Conservative Policy Iteration (CPI) is a founding algorithm of Approximate Dynamic Programming (ADP). Its core principle is to stabilize greediness through stochastic mixtures of consecutive policies. It comes with strong theoretical guarantees, and inspired approaches in deep Reinforcement Learning (RL). However, CPI itself has rarely been implemented, never with neural networks, and only experimented on toy problems. In this paper, we show how CPI can be practically combined with deep RL with discrete actions, in an off-policy manner. We also introduce adaptive mixture rates inspired by the theory. We experiment thoroughly the resulting  algorithm on the simple Cartpole problem, and validate the proposed method on a representative subset of Atari games. Overall, this work suggests that revisiting classic ADP may lead to improved and more stable deep RL algorithms.
\end{abstract}

\section{Introduction}

We consider the Reinforcement Learning (RL) problem with discrete actions, formalized with Markov Decision Processes (MDP)~\cite{puterman1994markov}. Approximate Dynamic Programming (ADP) is a standard approach to practically solve MDPs when the state space is large. In this case, a popular~-- and rather successful~-- approach is to approximate the value function and/or the policy with function approximation, using techniques ranging from linear parametrization to deep neural networks. Recently, several algorithms inspired by ADP have shown unprecedented results on hard control tasks by using deep neural networks, that provide a great power of approximation. A lot of these algorithms can be seen as instances or variations of ADP algorithms, notably Value Iteration (VI) and Policy Iteration (PI). For example, Deep Q-Network (DQN)~\cite{mnih2015human} can be related to VI, while Soft Actor-Critic (SAC)~\cite{haarnoja2018soft} or Trust Region Policy Optimization (TRPO)~\cite{schulman2015trust} can be related to PI.

Conservative Policy Iteration (CPI) is a classic extension of PI introduced by~\citet{kakade2002approximately}. Its main principle is to relax the improvement step in PI by being conservative with respect to the previous policies: instead of computing a sequence of deterministic greedy policies (as in PI), CPI computes a sequence of stochastic policies that are mixtures between consecutive greedy policies. While CPI has inspired some recent algorithms, such as TRPO~\cite{schulman2015trust}, it has never been implemented as such in practice, nor experimented on large challenging environments. In this paper, we propose a way to derive a practical algorithm from CPI, using neural networks as approximation scheme and relaxing the on-policy nature of CPI into off-policy learning through a VI-like scheme. We call the resulting algorithm Deep Conservative Policy Iteration, or DCPI (even if it is VI-based, to highlight the connection to CPI). It is specifically a conservative variation of DQN, but the proposed approach could be in principle applied to any pure-critic algorithm, notably the many variations of DQN.

After a short background, we develop the approximation steps that allow us to go from CPI to DCPI (Sec.~\ref{relax}), and give a detailed description of DCPI (Sec.~\ref{dcpi}). We then discuss some adaptive mixture rates in Sec.~\ref{choosing}, inspired by the theory, and present experimental results on Cartpole and Atari environments in Sec.~\ref{experiments}.  %

\section{Background and notations}

We classically frame RL with an infinite horizon discounted MDP, a tuple $\{\states, \actions, P, r, \gamma\}$ where $\states$ is the state space\footnote{We assume it finite, for the ease of notations, but what we present extends to the continuous case.}, $\actions$ the finite action space, $P \in \Delta_\states^{\states \times \actions}$ the Markovian transition kernel, $r \in [-R, R]^{\states \times \actions}$ a bounded reward function, and $\gamma \in (0,1)$ a discount factor. A stochastic policy $\pi$ associates to each state $s$ a distribution over actions $\pi(\cdot | s)$. We write $P_\pi(\p s | s) = \mathbb{E}_{a \sim \pi(\cdot | s)}[P(\p{s}|s,a)]$ for the stochastic kernel associated to $\pi$, and $r_\pi(s) = \mathbb{E}_{a \sim \pi(\cdot | s)}[r(s,a)]$ the expected discounting reward for starting in $s$ and following $\pi$. The value $v_\pi \in \mathbb{R}^{\states}$ of a policy is, for all $s \in \states$,
\begin{equation*}
    v_\pi(s) = \mathbb{E}_{\pi}\left[\sum_{t=0}^{\infty} \gamma^t r(s_t, a_t) \,\middle\vert\, s_0=s \right],  %
\end{equation*}
where $\mathbb{E}_{\pi}$ designates the expected value over all trajectories produced by $\pi$. The value function of a policy is the unique fixed point of the Bellman evaluation operator associated to this policy, defined for each $v \in \mathbb{R}^{\states}$ as $T_\pi v = r_\pi + \gamma P_\pi v$. From this operator, one can define the Bellman optimality operator for each $v \in \mathbb{R}^\states$, $T_\star v = \max_\pi T_\pi v$. $T_\star$ admits as its unique fixed point  the optimal value $v_\star$. A policy is said to be greedy w.r.t. to a value function $v$ if $T_\pi v = T_\star v$, the set of all such policies is written $\gr v$. A policy $\pi_\star$ is optimal with value $v_{\pi_\star} = v_\star$ when $\pi_\star \in \gr v_\star$.
To any policy $\pi$, we also associate the quality function $q_\pi$, for each $(s,a) \in \states \times \actions$
\begin{equation*}
    q_\pi(s,a) = r(s,a) + \mathbb{E}_{\p{s} \sim P(\cdot|s,a)}[v_\pi(\p{s})],
\end{equation*}
which behaves similarly to the value function in the sense that $T_\pi q_\pi = q_\pi$ and $T_\star q_{\pi_\star} = q_{\pi_\star} = q_\star$ (with a slight abuse of notation). We can also define the set of policies that are greedy w.r.t. any function $q \in \mathbb{R}^{\states\times\actions}$ that we write $\gr q = \argmax_a q(\cdot,a)$. It is useful in practice because a policy can be greedy to a $q$-function even if the model (the transition kernel) is unknown.

Finally, the advantage of a policy $\pi$, $A_\pi$, is defined as $A_\pi(s,a) = q_\pi(s,a) - v_\pi(s)$, and we write $d_{\pi,\mu} = (1 -\gamma)\mu(I-\gamma P_{\pi})^{-1}$ the discounted cumulative occupancy measure induced by $\pi$ when starting from a distribution $\mu$ of states (distributions being written as row vectors).

\section{Relaxing CPI}
\label{relax}
In this section, we describe the process that leads from CPI, a mainly theoretical dynamic programming algorithm, to a variant that can be combined with deep networks in an off-policy manner.
\subsection{Ideal CPI}
We first turn to the description of the CPI algorithm. We start by introducing the classic Approximate Policy Iteration (API)~\cite{bertsekas1996neuro}, an iterative scheme that takes as input a distribution $\mu$ of states, and that computes at each iteration $k$ a new policy
\begin{equation} 
    \pi_{k+1} = \gr q_k,
\end{equation}
where $q_k$ is an approximation of $q_{\pi_k}$ computed with states sampled from $\mu$. An error on the greedy step $\gr$ can be considered, but this error only appears when considering a infinite action space or when the greedy policy is approximated (for example with a cost-sensitive classifier). Here, we consider a finite action space, the greediness with respect to a $q$-function is exact.

CPI was first proposed by~\citet{kakade2002approximately}. At each iteration $k$, CPI uses a mixture coefficient $\alpha_k$ to compute a stochastic mixture of all the previous greedy policies,
\begin{equation}
    \label{cpi}
    \pi_{k+1} = (1 - \alpha_{k+1})\pi_k  + \alpha_{k+1}\gr q_k,
\end{equation}
where $q_k$ is still an approximation of $q_{\pi_k}$. This algorithm comes with strong theoretical guarantees, in particular the mixture rate can be chosen so that Eq.~\eqref{cpi} guarantees an improvement of the expected value of the policy value, as shown by~\citet{kakade2002approximately} and~\citet{pirotta2013safe}. In these works, the error on the value function estimation is supposed bounded, and the mixture rate depends on this bound. These theoretical guarantees rely on the fact that at each iteration $k$, the approximations are computed on the distribution $d_{\pi_k, \mu}$, where $\mu$ is the starting distribution of states, something far from being practical and making CPI inherently on-policy. CPI and its extension Safe Policy Iteration (SPI)~\cite{pirotta2013safe} have only been experimented on tabular toy problems, with at most linear function approximation, in a very controlled manner~\cite{pirotta2013safe,scherrer2014approximate}.

We will next introduce approximations that allow for an actual implementation using deep learning in an off-policy setting, but keeping the essence of CPI, that is regularizing the greediness. The question of the choice of the mixture rate will be studied later. %

\subsection{Approximating towards practicality}

\paragraph{Approximating the value} First, as said before, the value function has to be approximated. As the distributions $d_{\pi_k, \mu}$ are impractical, one classically computes an estimate $q_k$ of the quality function $q_{\pi_k}$, with states sampled from a fixed state distribution or gathered during learning. The quality function can be estimated either by rollouts -- but this  is quite sample inefficient -- or for example by using an algorithm such as LSTD~\cite{bradtke1996linear} -- but that would require a linear parametrization. In any case, we can consider an error $\epsilon_k$ on this approximation, resulting in the scheme
\begin{equation}
\label{cpi_2}
\begin{cases}
    q_{k} = q_{\pi_{k}} + \epsilon_{k}\\
    \pi_{k+1} = (1 - \alpha_{k+1})\pi_k  + \alpha_{k+1}\gr q_k.\\
\end{cases}
\end{equation}
\paragraph{Temporal differences} A classic approach is Temporal Difference (TD) learning, that estimates $q_k(s,a)$ by performing a regression on targets of the form $r(s,a) + \gamma \sum_{\p a\in\actions} \pi_k(\p a|\p s) q_{k-1}(\p s,\p a)$. This can be written formally as computing $q_{k+1} = T_{\pi_k}q_{k-1} + \epsilon_k$. Practically, one can consider doing m-steps returns~\cite{sutton1988learning}, which from an abstract perspective is $q_k = T_{\pi_k}^m q_{k-1}$, as done in Modified Policy Iteration (MPI)~\cite{puterman1978modified}, or even Approximate MPI~\cite{scherrer2015approximate}. This results in the scheme
\begin{equation}
\label{cpi_3}
\begin{cases}
    q_{k} = T_{\pi_{k}}^m q_{k-1} + \epsilon_{k}\\
    \pi_{k+1} = (1 - \alpha_{k+1})\pi_k  + \alpha_{k+1}\gr q_k.\\
\end{cases}
\end{equation}
Note that with $m=\infty$, it falls back to Eq.~\eqref{cpi_2}, and with $m=1$, it becomes similar to VI, where the greediness has been regularized. Specifically, with $m=1$ and $\alpha_k = 1$, this reduces to AVI (Approximate VI).
In addition to allow using TD-learning, this also allows to learn in an off-policy manner (without off-policy correction if $m=1$, as we work with state-action value functions).

\paragraph{Approximating the mixture} Computing $\pi_k$ would require remembering every $q_i$ computed for $i\in[0,k]$, and this is not feasible in practice. Instead, we approximate the mixture, which adds a new source of errors. This can be done, for example, by minimizing an expected Kullback-Leibler (KL) divergence between a parametrization of $\pi_{k+1}$ and the mixture. It can be written formally as
\begin{equation}
\label{cpi_4}
\begin{cases}
    q_{k} = T_{\pi_{k}}^m q_{k-1} + \epsilon_{k}\\
    \pi_{k+1} = (1 - \alpha_{k+1})\pi_k  + \alpha_{k+1}\gr q_k + \p{\epsilon_{k+1}}.
\end{cases}
\end{equation}
This approximate dynamic programming scheme can then be instantiated into an off-policy Deep RL algorithm; we detail this process in Section~\ref{dcpi}.

\subsection{Theoretical insights}

The scheme depicted on Eq.~\eqref{cpi_4} no longer enjoys the theoretical guarantees of CPI, as we relax some of its components (for example, partial policy evaluation or more freedom on how samples are gathered for learning). We give a partial analysis of this relaxed scheme in the Appendix, and here, we discuss its main results. Without errors ($\epsilon_k = \epsilon'_k = 0$),  we show in the Appendix that $v_{\pi_k}$ will converges linearly to $v_*$. With $\alpha_k=1$ (this corresponds to MPI), the scheme benefits from a $\gamma$-contraction and it leads to a bound $\|v_* - v_{\pi_k}\|_\infty = \mathcal{O}(\gamma^k)$. With $\alpha_k<1$, we obtain an $\eta_k$-contraction with $\eta_k=1-\alpha_k(1-\gamma)$. If $\alpha_k$ does not go too fast towards zero, this would also lead to linear convergence. Indeed, using the fact that $\ln(1-x) \leq -x$ for $x\in(0,1)$,
\begin{equation}
\begin{split}
    \prod_{i=1}^k \eta_k & = \exp \sum_{i=1}^k \ln(1 - \alpha_i(1-\gamma))\\
                         & \leq \exp(-(1-\gamma)\sum_{i=1}^k \alpha_i).\\
\end{split}
\end{equation}
Therefore, this would lead to a bound $\|v_* - v_{\pi_k}\|_\infty = \mathcal{O}(\prod_{i=1}^k \eta_i) = \mathcal{O}(\exp(-(1-\gamma)\sum_{i=1}^k \alpha_i))$. If we still have a linear convergence, it is slower as long as $\alpha_k<1$, which was to be expected without approximation error. However, at least this scheme does not break convergence.

With errors, we conjecture that we would obtain a bound close to the one of AMPI~\cite[Thm.~7]{scherrer2015approximate}, maybe with a larger propagation of errors (much like the convergence is slower, in the exact case), and so worse than the original bound of CPI~\cite{kakade2002approximately,scherrer2014approximate} (notably, with bigger concentrability coefficient). This is to be expected, the bound of CPI relies heavily on using $m=\infty$, on how the approximation error is plugged in the approximate dynamic scheme, and on using the $d_{\pi,\mu}$ distribution to sample transitions for learning approximations, three things that we relax. Yet, we still think that relaxing greediness is worth experimentally speaking, and that much remains to be done regarding its theoretical understanding.

\section{Deep CPI \label{dcpi}}

We now turn to the actual practical algorithm, DCPI. The basic idea is to define an instance of the update in Eq.~\eqref{cpi_4} where the value function and the policy are parametrized via neural networks. We will focus on the case $m=1$ (a regularized VI scheme), so we can apply the evaluation operator to the estimated $q$-function in an off-policy fashion without correction. It could be extended to the case $m>1$ by simply using an off-policy correction method such as importance sampling. Note that focusing on $m=1$ makes our algorithm a regularized VI-scheme and not a PI-scheme, but we keep the name DCPI to highlight the connection to CPI.

We parametrize the $q$-function and the policy by two \emph{online} networks $q_\theta$ and $\pi_\omega$, where $\theta$ and $\omega$ denote the weights of the respective networks. In a similar way to DQN, we define two \emph{target} networks, $\old q$ and $\old \pi$, whose weights are respectively $\old\theta$ and $\old\omega$. DCPI introduces stochastic approximation by acting in an online way, meaning that transitions $(s,a,r,\p{s}) \in \states  \times \actions \times \mathbb{R} \times \states $ from the environment are collected during training. Transitions are stored in a FIFO replay buffer $\mathcal{B}$. 

We  write the two updates from Eq.~\eqref{cpi_4} as optimization problems. The evaluation step consists of a regression problem, trying to minimize a quadratic error between $q_\theta$ and an approximation of $T_{\pi_\omega}^{m} \old q$. Recall that we now use $m=1$. From this, denoting $\avg$ the empirical mean over a finite set, we can  define a regression loss function $\mathcal{L}_q(\theta)$ for the value weights as 
\begin{equation} \label{qtarget}
    \avg \left[\left(r + \gamma \sum_{\p{a} \in \actions} \old\pi(\p{a} \vert \p{s})\old q(\p{s}, \p{a}) - q_{\theta}(s, a)\right)^2\right],
\end{equation}
where the empirical average is computed over all transitions $(s,a,r,\p{s}) \in \mathcal{B}$ (recall that there is no need for these transitions to be sampled according to $\pi^{-}$, as we learn off-policy). The improvement step requires approximating a distribution over actions for each state. One way to do that is to minimize the expected value over the states of the expected KL divergence between the online policy network and the stochastic mixture. This leads to a loss function $\mathcal{L}_\pi(\omega)$ on the policy weights,
\begin{equation} \label{ptarget}
    \avg \left[\kl\left((1 - \alpha) \old\pi(\cdot \vert s) + \alpha \gr(q_{\theta}) (\cdot | s) \Vert \pi_{\omega}(\cdot \vert s)\right)\right],
\end{equation}
where the empirical average is computed over all states $(s, \dots)\in \mathcal{B}$. We minimize both $\mathcal{L}_q$ and $\mathcal{L}_\pi$ with a fixed number of steps of batch-SGD (or a variant), and update the target networks with the weights of the online networks. Each gradient step is performed after a fixed number (the interaction period $F$) of transitions are collected from the environment. Note that the use of a replay buffer makes our algorithm off-policy: the samples used to evaluate $\pi_w$ originate independently from older policies.  During training we sample transitions with $\pi_{\omega,\varepsilon}$, the policy which chooses a random action uniformly on $\actions$ with probability $\varepsilon$ and follows $\pi_\omega$ with probability $1-\varepsilon$ (recall that $\pi_\omega$ is itself stochastic). A detailed pseudo-code is given in Algorithm~\ref{algo_dcpi}.

\paragraph{Connection to DQN} Despite its actor-critic look, DCPI can simply be seen as a variation of DQN. Indeed, note that with $\alpha=1$, if $\pi_\omega$ is exactly computed (\textit{i.e.} if $\pi_w = \gr q_\theta$), DCPI reduces to DQN.

\begin{algorithm}
\caption{DCPI}
\label{algo_dcpi}
\begin{algorithmic}[1]
\REQUIRE $K\in \mathbb{N^\star}$ the number of steps, $C\in \mathbb{N^\star}$ the update period, $F \in \mathbb{N^\star}$ the interaction period
\STATE Initialize $\theta$, $\omega$ at random
\STATE $\mathcal{B} = \{\}$
\STATE $\old\theta = \theta, \old\omega = \omega$
\FOR{$k = 1$ \TO $K$}
    \STATE Collect a transition $t = (s, a, r, \p{s})$ from $\pi_{\omega, \varepsilon}$
    \STATE $\mathcal{B} \leftarrow \mathcal{B} \cup \{t\}$
    \IF{$k \mod F == 0$}
        \STATE On a random batch of transitions $B_{q,k} \subset \mathcal{B}$, update $\theta$ with one step of SGD of $\mathcal{L}_q$, see~\eqref{qtarget} \label{line:lq}
        \STATE On a random batch of transitions $B_{\pi,k} \subset \mathcal{B}$,
            update $\omega$ with one step of SGD of  $\mathcal{L}_\pi$, see~\eqref{ptarget} \label{line:lpi}
    \ENDIF
    \IF{$k \mod C == 0$}
        \STATE $\old\omega \leftarrow \omega$, $\old\theta \leftarrow \theta$
    \ENDIF
\ENDFOR
\RETURN $\pi_{\omega}$
\end{algorithmic}
\end{algorithm}

\section{Choosing the mixture rate \label{choosing}}

Algorithm~\ref{algo_dcpi} does not give a way to choose the mixture rate, and this section studies different manners to do it. The natural idea is to choose a constant rate which experimentally (see Section~\ref{cartpole}) seems to improve stability, but comes at a great cost in terms of sample efficiency. Another possibility is to choose a decaying rate, for example with a hyperbolic schedule, or -- and that is what we focus on -- choosing an adaptive rate inspired from the literature on CPI.

\paragraph{CPI adaptive rate} 
\citet{kakade2002approximately} provide a rate for CPI that guarantees an improvement of the policies, by choosing $\alpha = \frac{(1-\gamma) \advmu}{4R}$. Here, we write $\bar\pi = \gr q_\pi$ the greedy policy with respect to $q_{\pi}$, and $\advmu$  the advantage of the greedy policy ($\bar\pi$) over the previous one ($\pi$), that is   $\advmu = \sum_{s \in \states}d_{\pi, \mu}(s)\adv(s)$ with $\adv(s) = \sum_{a \in \actions} \bar\pi(a|s) A_{\pi}(s,a)$. Recall that $R$ is the maximum possible reward. We can estimate close quantities over a batch $B \subset \mathcal{B}$ at step $k$ in the sense of Algorithm~\ref{cpi}. We compute $\hat{A}_k(s) = \max_{a \in \actions} q_{\theta}(s,a) - \sum_{a \in\actions} \pi_\omega(a|s)q_\theta(s,a)$ as an estimate of $\adv(s)$, and $\hat{\mathbb{A}}_k = \avg_{(s, \hdots) \in B}[\hat{A}_k(s)]$ as an estimate of $\advmu$. The term $R/(1-\gamma)$ can be approximated by an estimate $\hat Q_k$ of $\Vert q_\pi \Vert_\infty$, which is consistent with corollary 3.6 of~\citet{pirotta2013safe}. We compute it over a batch with $\hat Q_k = \max_{(s,a,\hdots) \in B} |q_\theta(s,a)|$. For simplicity and to add a degree of freedom, we replace the constant factor $1/4$ by an hyperparameter $\alpha_0$ that allows us to directly control the amplitude of our mixture rate. To compensate the fact that we compute our approximation over (potentially small) batches, we use a moving average $m_k$ and a moving maximum $Q^{+}_k$. This leads to
\begin{equation}
\label{alpha_cpi}
\begin{cases}
    m_k = \beta_1 {m_{k-1}} + (1 - \beta_1) \hat{\mathbb{A}}_{k}\\
    Q_k^{+} = \max{(\beta_2 Q_{k-1}^{+}, \hat{Q}_k)}\\
\end{cases}, \quad
     \alpha_{k}^{cpi} = \alpha_0 \frac{m_k}{Q_k^{+}},
\end{equation}
with $\beta_1, \beta_2 \in (0,1)$ typically close to $1$.

\paragraph{SPI adaptive rate} 
\citet{pirotta2013safe} propose an improvement of CPI, Safe Policy Iteration. They provide a better bound on the policy improvement based on the mixture rate $\alpha = \frac{(1 - \gamma)^2 A_{\pi,\mu}^{\bar{\pi}}}{\gamma \| \bar{\pi} -  \pi\|_\infty \Delta \adv}$, with $\Delta \adv = \max_{s \in \states}\adv(s) - \min_{s \in \states}\adv(s)$, and with $\| \bar{\pi} -  \pi\|_\infty = \max_{s\in\states}\sum_a|\pi(a|s) - \old\pi(a|s)|$ the maximum total variation between policies. We can approximate these quantities with the same methods used to obtain Eq.~\eqref{alpha_cpi}. Using the value $\hat{A}_k$ described previously, we compute an estimate of $\Delta \adv$ by subtracting $\hat{A}_{k,\min} = \min_{(s,\hdots) \in B}\hat{A}_k(s)$ to $\hat{A}_{k,\max} = \max_{(s,\hdots) \in B}\hat{A}_k(s)$. Note that in addition to the previous approximations, we also include the total policy variation in the $\alpha_0$ hyperparameter, as $\| \bar{\pi} -  \pi\|_\infty \leq 2$. Using moving approximations, we obtain
\begin{equation} \label{alpha_spi}
\begin{split}
\begin{cases}
     M_k^{+} = \max{(\beta_2 M_{k-1}^{+}, \hat{A}_{k,\max})}\\
     M_k^{-} = \min{(M_{k-1}^{-}/\beta_2, \hat{A}_{k,\min})}\\
\end{cases}&, \\
\alpha_{k}^{spi} = \alpha_0 \frac{m_k}{M_k^{+} - M_k^{-}}.&
\end{split}
\end{equation}

\paragraph{Bounding SPI} 
The SPI mixture rate from Eq.~\eqref{alpha_spi} gives a rate that is not bounded. To keep our rate below~$1$, we propose a simple variation
\begin{equation}
\label{alpha_adx}
\alpha_{k}^{adx} = \alpha_0 \frac{m_k}{M_k^{+}}.
\end{equation}
From the fact that $\hat{A}_k(s)$ are positive numbers, it is immediate that $\alpha^{adx}$ is a ``little more conservative'' version of $\alpha^{spi}$, with $\alpha^{adx} \leq \alpha^{spi}$ and $\alpha^{adx} \leq 1$. In fact, the advantage function can be linked to the functional gradient of the expected value function, respectively to the policy (see~\citet{scherrer2014local} who interpret CPI as a policy gradient boosting approach) and this rate is similar to the one the Adamax~\cite{kingma2014adam} algorithm would give (up to the fact that our rate is global, not component-wise) -- hence the name. 

\paragraph{About the batch}
The adaptive rate is computed using a batch of transitions from the replay buffer, and an important question is \emph{which} batch to choose. In Algorithm~\ref{algo_dcpi}, two different batches of transitions are defined: $B_{q,k}$ a batch of transitions used to estimate $q_\theta$, and $B_{\pi,k}$ used to estimate $\pi_\omega$. Our approach is, as the rate needs to adapt with respect to the current policy, to use $B_{\pi,k}$ to compute the rate. That means that, at iteration $k$ in Algorithm~\ref{algo_dcpi}, $\alpha_k$ and $\hat \nabla_{\omega} \mathcal{L_\pi}$ (the approximation of the gradient of $\mathcal{L}_\pi$ computed at line~\ref{line:lpi} of Algorithm~\ref{algo_dcpi}) are computed with the same batch of transitions.

\section{Experiments \label{experiments}}
In this section, we experimentally study DCPI on several environments. The method we propose is general, and could be used to regularize any pure-critic algorithm, by adding an actor to  it. For this experimentation, we consider DCPI as a variation of DQN, and take DQN as our baseline. In principle, our method could extend to other frameworks, such as Rainbow~\cite{hessel2018rainbow} or Implicit Quantile Networks (IQN)~\cite{dabney2018implicit}, which are extensions to DQN. We start this experiment by an intensive test on Cartpole, a light environment that allows us to exhibit various behaviours of DCPI, such as stability over random seeds, convergence speed, or efficiency of the proposed mixture rate. We then conduct an experiment on Atari, to observe the effects of scaling up.

\subsection{Cartpole \label{cartpole}}
Cartpole is a classic control problem introduced by \citet{barto1983neuronlike}. In this setup, the agent needs to balance a vertical pole by controlling its base (the cart) along one dimension, by applying a force on the cart of $-1$ or $+1$. We use the version of Cartpole implemented in OpenAI Gym~\cite{brockman2016openai}, with a maximum steps limit raised to $500$ steps instead of a more classic $200$, to make the task harder and get more accurate observations. The agent gets a reward of $+1$ while the pole is in the air, and $0$ when it touches the ground. 

Although CartPole is considered an ``easy'' problem in RL, it is cheap to run in computation time, so we use it as a test-bed to perform studies on the influence of our hyperparameters. Such studies would be prohibitive in cost on larger environments such as Atari. Our approach is to modify the DQN algorithm without changing its parameters so as to analyze how our framework modifies its learning behaviour.
\begin{figure}[t]
	\begin{center}%
	\includegraphics[width=0.95\linewidth]{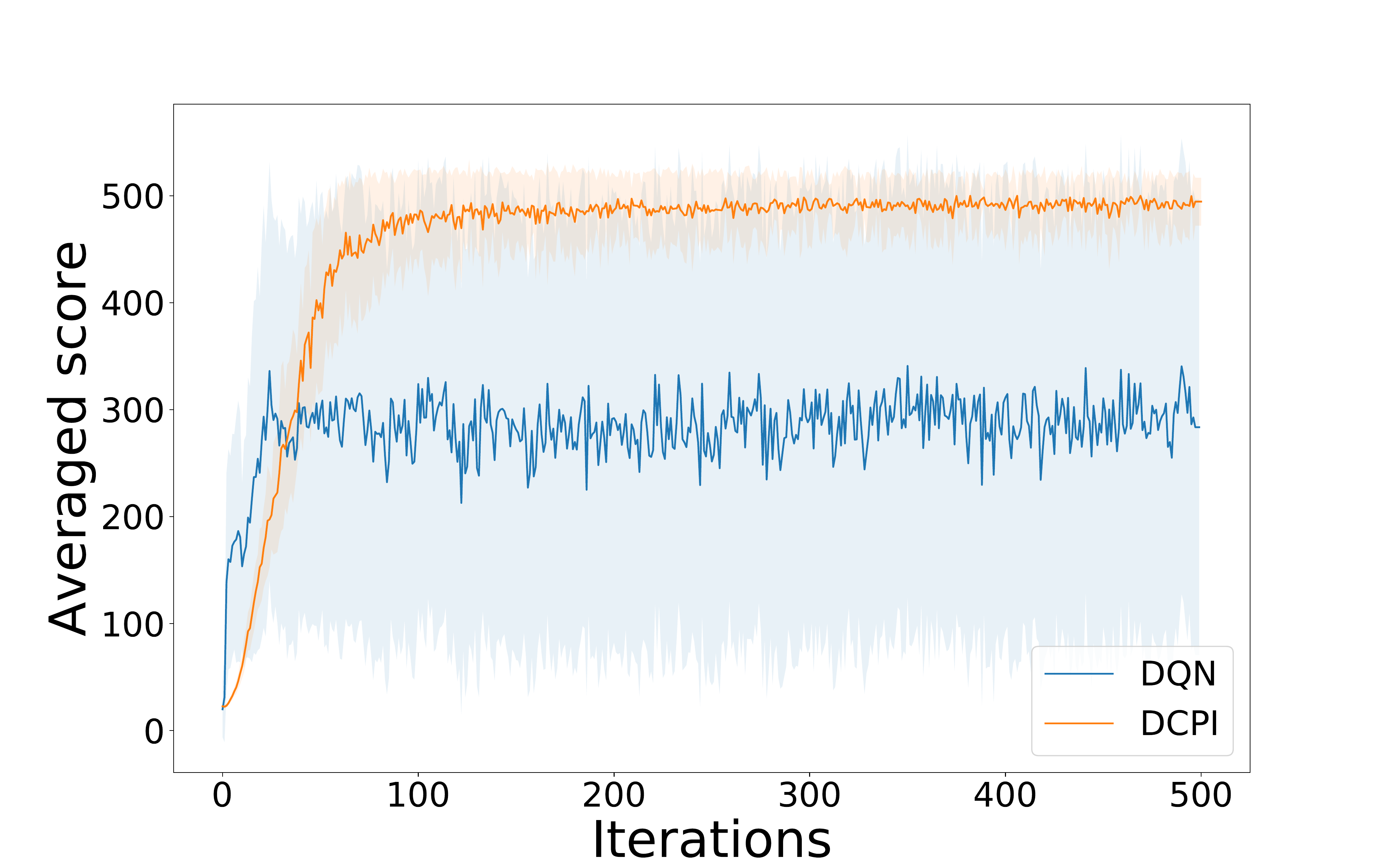}
	\\
	\includegraphics[width=0.95\linewidth]{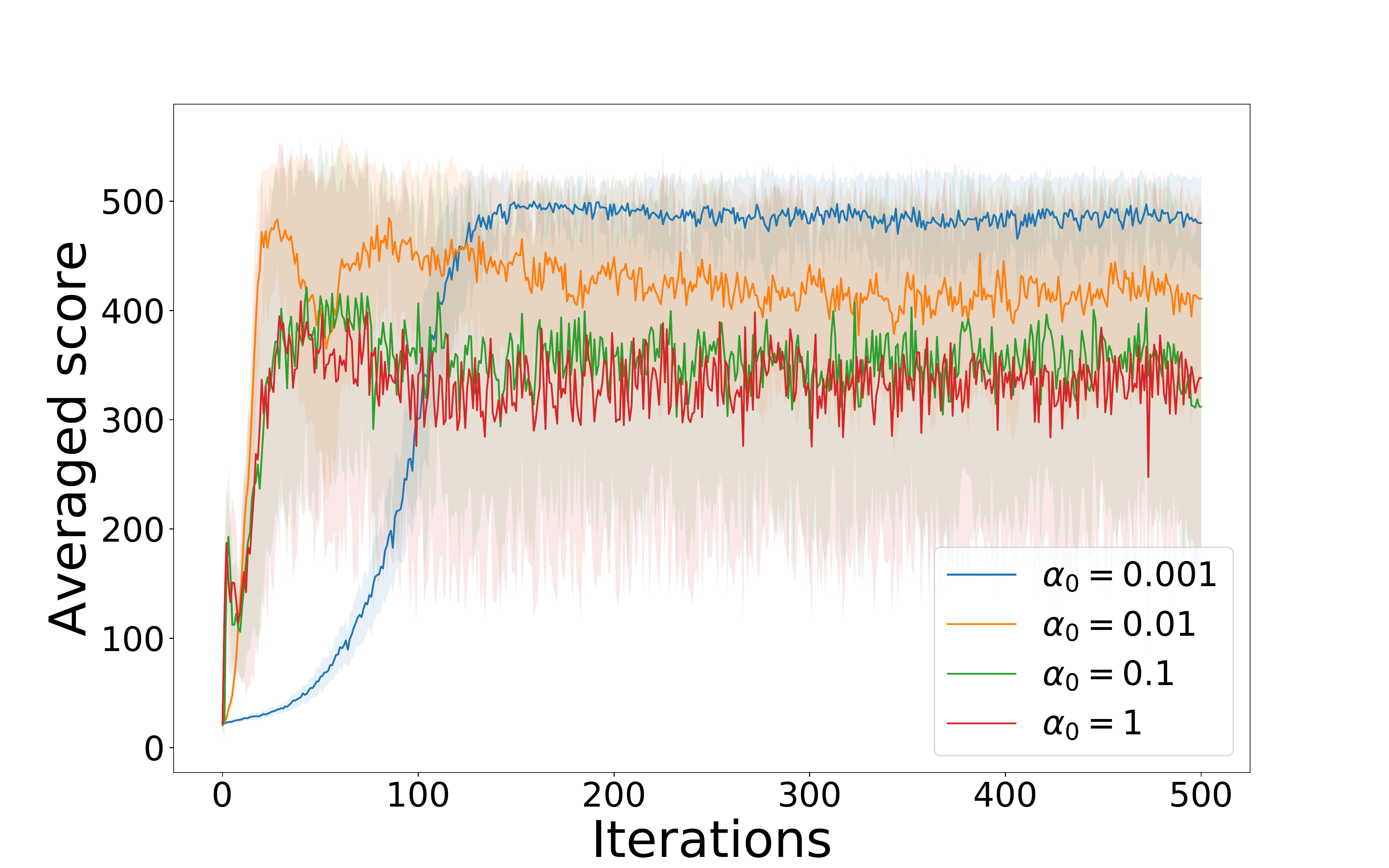}
	\end{center}%
	\caption{\textbf{Top:} comparison of the averaged  training scores of DCPI with CPI rate and $\alpha_0=0.1$ (orange) against DQN (blue). \textbf{Bottom:} DCPI on Cartpole with constant rates for $4$ values of $\alpha_0$.
	\label{dqn_cart} \label{fig:best} \label{fig:cst}}
\end{figure}
\begin{figure}[t]
    \centering
    \includegraphics[width=0.95\linewidth]{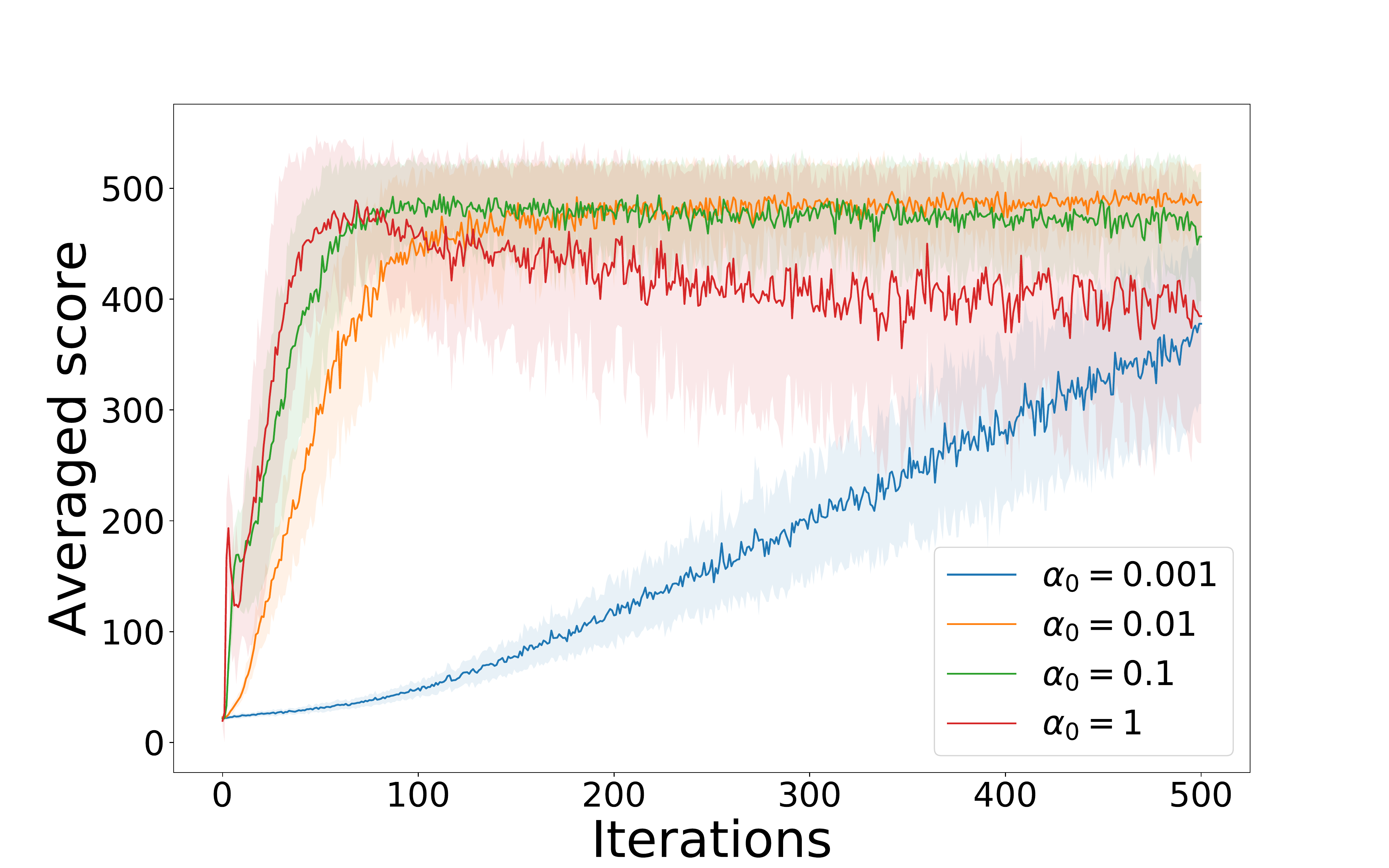}
    \\
    \includegraphics[width=0.95\linewidth]{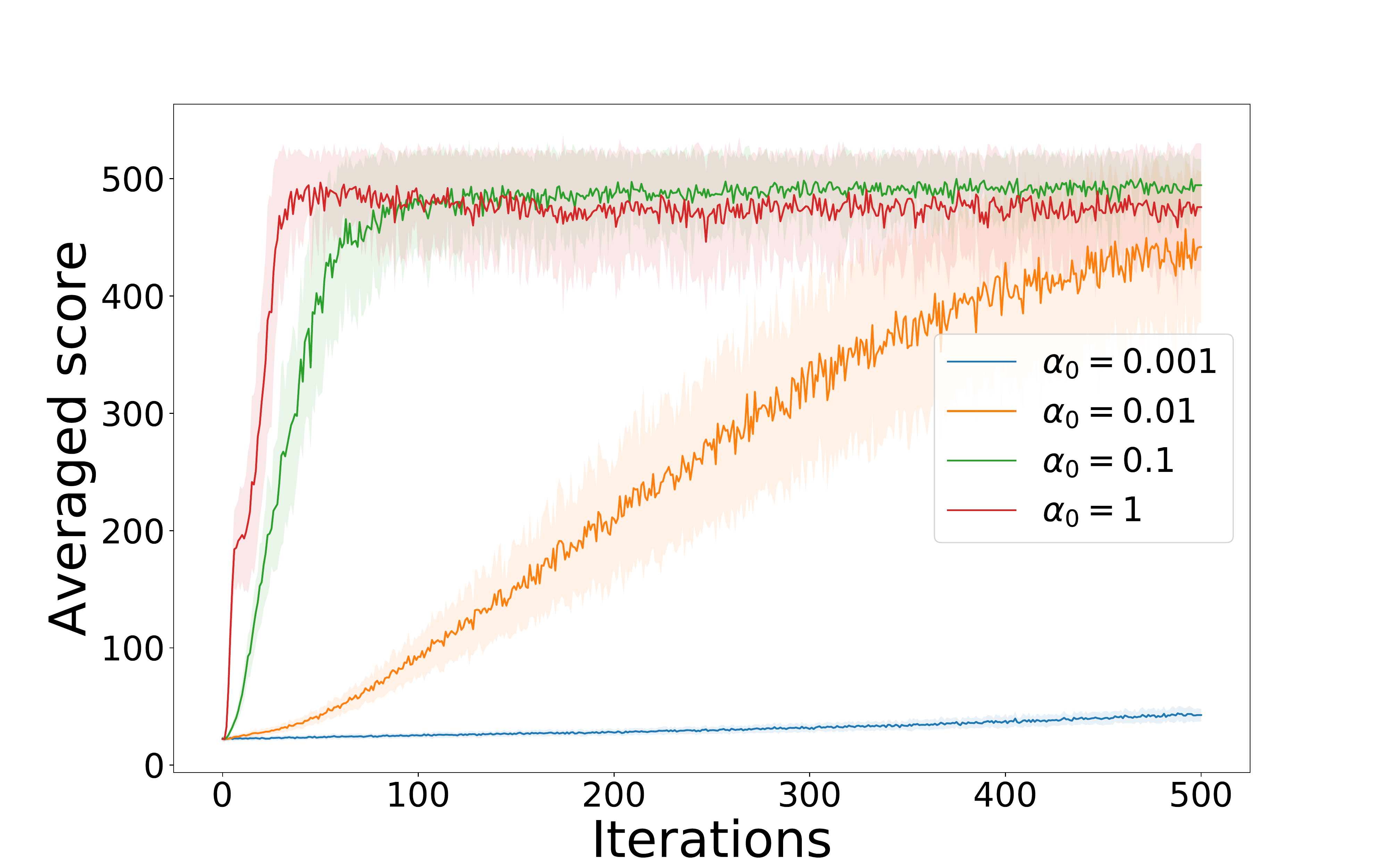}
    \caption{\textbf{Top:} DCPI on Cartpole with Adamax rates for $4$ values of $\alpha_0$. \textbf{Bottom:} DCPI on Cartpole with CPI rates for $4$ values of $\alpha_0$.
       \label{fig:adx} \label{fig:cpi}}
\end{figure}
Our baseline is  the DQN provided in the Dopamine library~\cite{castro2018dopamine}, and we use the hyperparameters provided here for Cartpole. Notably, we used the same network architecture for the q-network and the policy network and two identical Adam optimizers; we compute a gradient step every $F=4$ interactions with the environment, and update the target networks every $C=100$ interactions. Full parameters are reported in the Appendix. Our first observation is that this version of DQN is not very efficient on this problem, as it greatly lacks stability, be it over random seeds or over time (see Figure~\ref{dqn_cart}). This instability could probably be tempered by a better tuning of hyper parameters, but our goal is to verify the stabilizing effects of CPI, so we keep them as is.

Our method introduces three new hyperparameters: $\alpha_0, \beta_1,$ and $\beta_2$, described precisely in Section~\ref{choosing}. To choose $\beta_1$ and $\beta_2$, we consider that our estimate of the advantage should be stable between two updates of the target networks. As this update occurs every $100$ steps, and the size of the window for our moving average is $1/(1-\beta_1)$, this leads us to choose $\beta_1 = 0.99$. To increase stability, we choose a slower moving average in the denominator with $\beta_2=0.9999$. The ratio $(1-\beta_1)/(1-\beta_2)=100$ is classic, it is for example consistent with the defaults parameters of Adam~\cite{kingma2014adam}. We did a parameter search over $\alpha_0$, with values ranging from $1e-3$ to $1$, and tested the $\alpha^{cpi}$ and  $\alpha^{adx}$ heuristics for an adaptive rate described in Section~\ref{choosing}, Eqs.~\eqref{alpha_cpi} and~\eqref{alpha_adx}, in addition to a constant rate. The results for $\alpha^{spi}$ are similar to $\alpha^{adx}$, and provided in the Appendix.

Results presented in Figure~\ref{dqn_cart} and~\ref{fig:cpi} are computed as follows: every $1000$ training steps, an \emph{iteration} in this context, we report the averaged undiscounted score per episode over these $1000$ steps. The results are averaged over $50$ different random seeds: the thick line indicates the empirical mean, while the semi-transparent areas denote the standard deviation of the score over the seeds.

Results with a constant rate (see Figure~\ref{fig:cst}) show a strong increase of stability with small mixture rates ($\alpha_0=0.001$), with a cost in speed. With a higher learning rate, we obtain a faster convergence, but we loose stability. This introduces a speed/stability dilemma, and using adaptive rates (see Figure~\ref{fig:adx}) allows us to get the best of both worlds. In a good case -- CPI adaptive rate with $\alpha_0=0.1$, see Figure~\ref{fig:best} (top) -- we can keep the stability of the small constant mixture rates, while benefiting from a relatively fast convergence, and here DCPI shows a clear improvement on DQN on stability and average performance: DCPI is able to stabilize at an average score of $480$ (on a maximum of $500$) with a low standard deviation around $20$, while DQN stabilizes around $300$, with a standard deviation of approximately $200$.
Remarkably, even for $\alpha=1$ (see Figure~\ref{fig:cst}), \textit{i.e.} when the stochastic mixture is not conservative and the regularization only comes from approximating the greediness, DCPI yields a slight improvement on stability over DQN. This can be seen as the distillation of the greedy policy, and is here less effective than a mixture scheme.

\subsection{Atari \label{atari}}

Atari is a challenging discrete-actions control environment, introduced by~\citet{bellemare2013arcade} consisting of 57 games. We used sticky actions to introduce stochasticity as recommended by~\citet{machado2018revisiting}. In a similar way to our Cartpole experiments, we used the DQN implementation from the Dopamine library as our baseline, keeping the parameters given in this library -- much more optimized than the one for Cartpole. We compare against DQN's baseline score given in Dopamine. %
The parameters are detailed in the Appendix. In particular, the states stored in the replay buffer consist of stacks of $4$ consecutive observed frames. With the same arguments as in Section~\ref{cartpole} we chose $\beta_1=0.9999$ and $\beta_2=0.999999$. After a small hyperparameter search on a few games (Pong, Asterix and Space Invaders), we chose $\alpha_0 = 1$ and the Adamax mixture rate (see Eq.~\eqref{alpha_adx}). Provided results are computed in a similar manner to the ones from Cartpole, except that here, an \emph{iteration} represents $250000$ environment steps. The results are averaged over $5$ different random seeds.

For Atari, we also found empirically that interacting with the policy $\pi_{q,\varepsilon}$ that is $\varepsilon$-greedy with respect to $q_{\theta}$ improved performance over playing with $\pi_\omega$. This is taken into account in the provided results. This can be seen as an optimistic controller regarding the stochastic policy $\pi_\omega$, as both policies $\pi_\omega$ and $\pi_{q,\varepsilon}$ converge to the same behavior in the exact case. It can also be seen as a regularization of the Bellman optimality operator used in DQN, without changes to the way samples are gathered.

We tested DCPI on a representative subset of 20 Atari games, chosen from the categories described in~\cite[Appendix~A]{ostrovski2017count}, excluding the hardest exploration games with sparse rewards -- our algorithm has no ambition to help with exploration. All results are provided in the Appendix. DCPI yields a clear improvement on performance on a large majority of those games, outperforming DQN on $15$ games over $20$. Note that choosing a lower rate $\alpha_0$ could increase stability and final performance, but also lower convergence speed. We chose to use rather aggressive adaptive rates on Atari due to constraints on computing time. 

\begin{figure}
	\begin{center}%
	\includegraphics[width=0.95\linewidth]{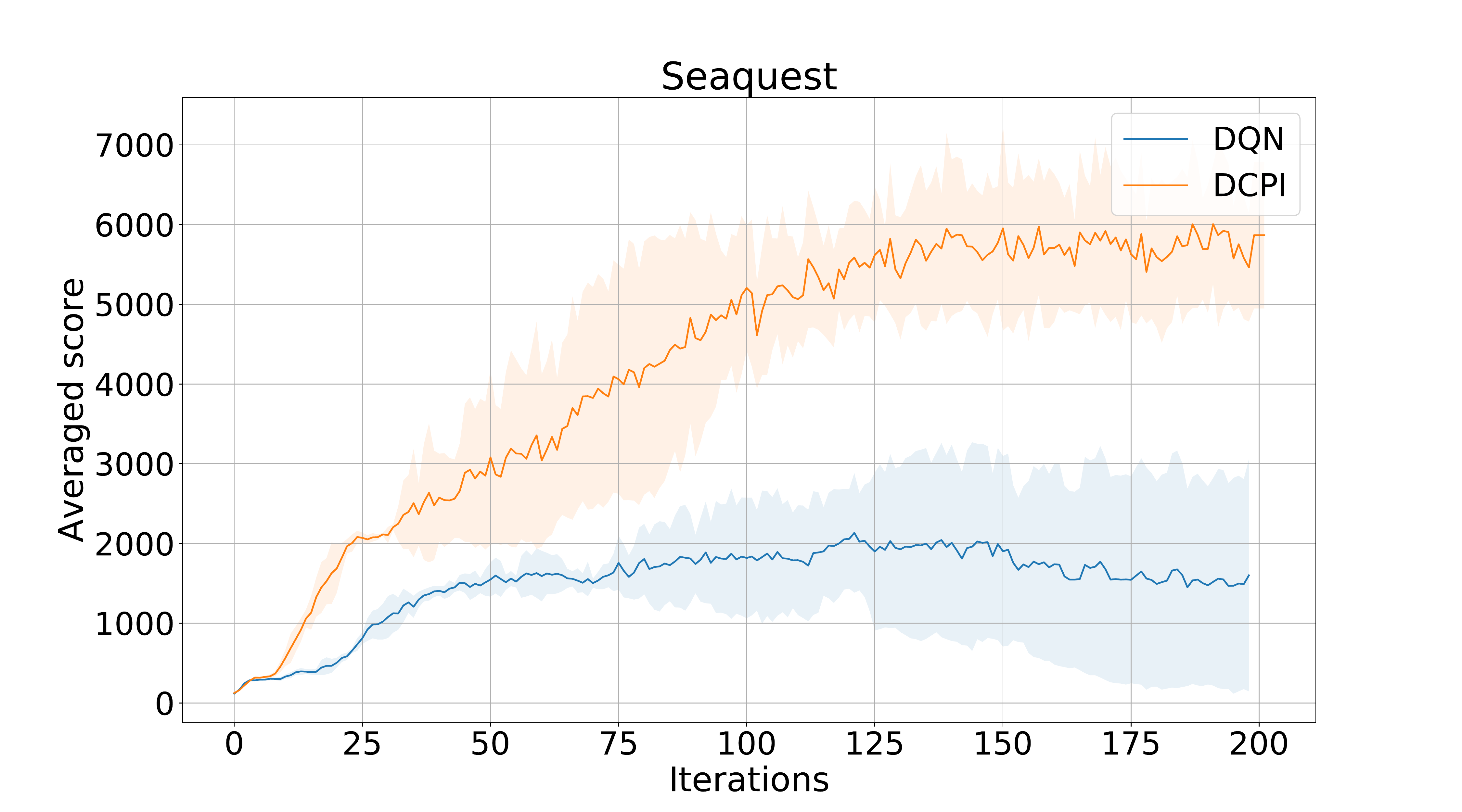}
	\\
	\includegraphics[width=0.95\linewidth]{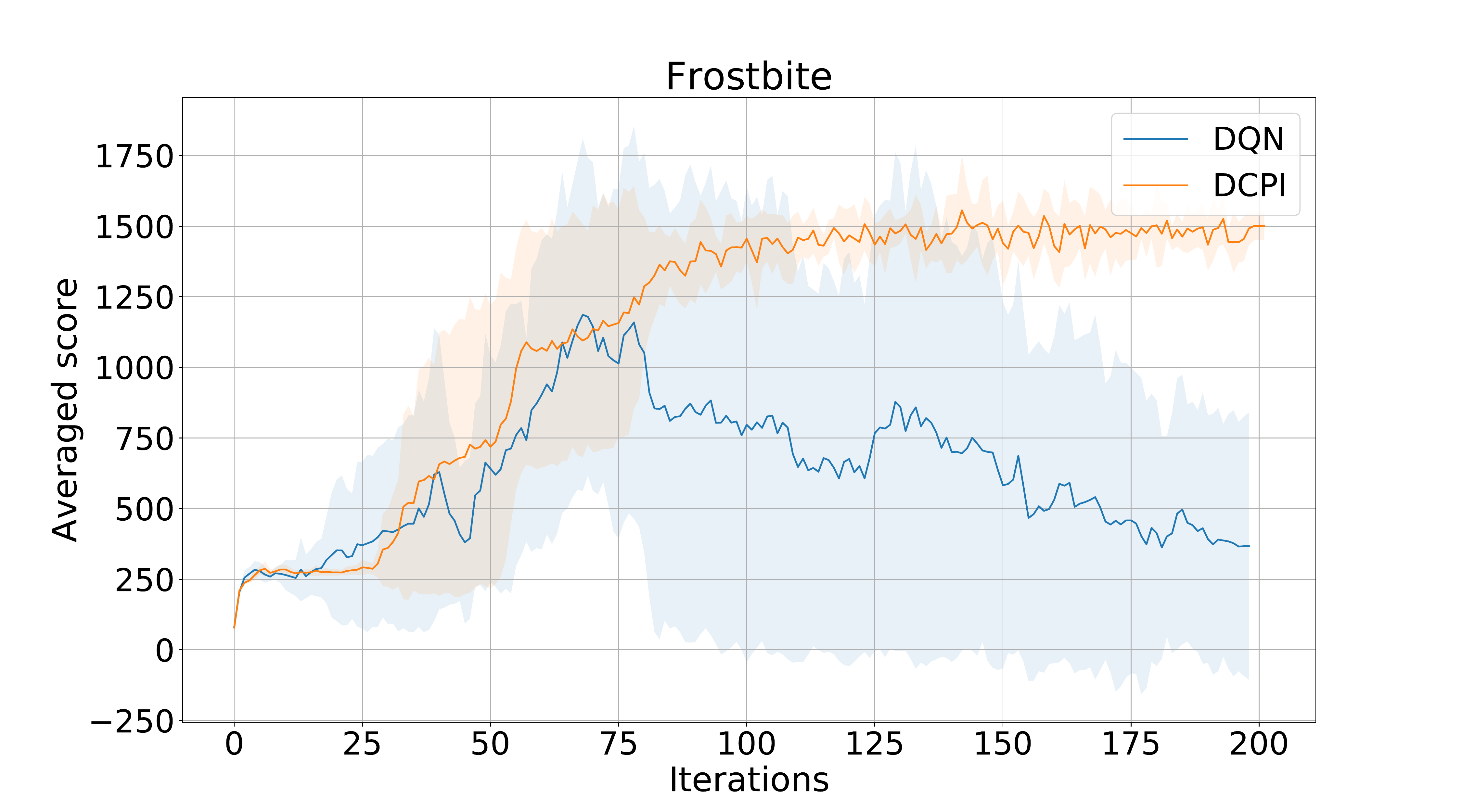}
	\\
	\includegraphics[width=0.95\linewidth]{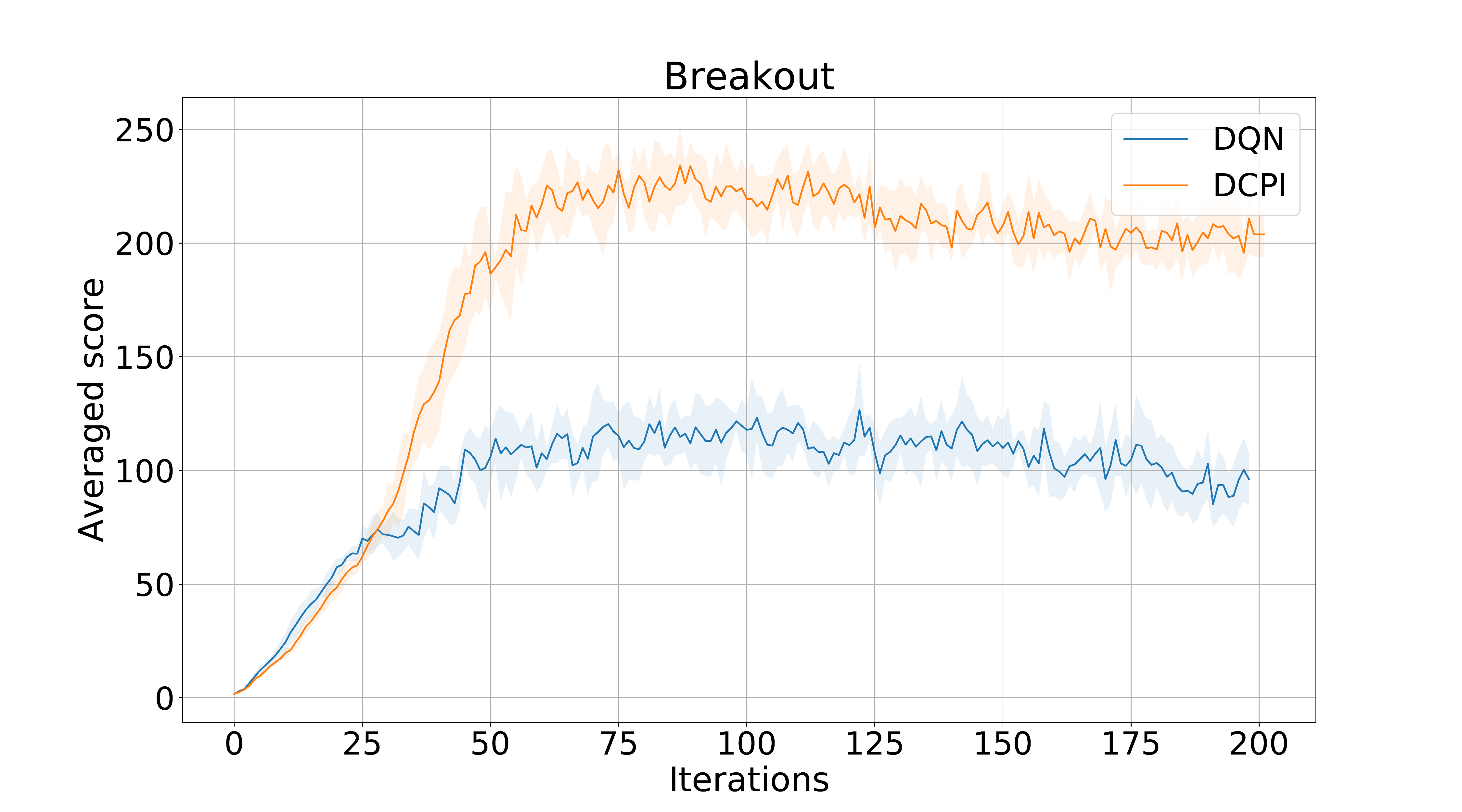}
	\end{center}%
	\caption{Averaged training scores of DCPI (orange) and DQN (blue) on three of the considered games (Seaquest, Frostbite and Breakout).
	\label{fig:ast} \label{fig:end}}
\end{figure}

As a matter of illustration, Figure~\ref{fig:ast} provides three games where DCPI attains a higher score than DQN: Seaquest, Frostbite, and Breakout. All other games are reported in the Appendix. We also report on Figure~\ref{fig:imp} a comparison summary of DQN vs DCPI on all considered games. We used the Area Under the Curve (AUC) metric. For each game, we compute the sum of all averaged returns obtained during training, respectively $S_{dcpi}$ and $S_{dqn}$, and we report the values for $(S_{dcpi} - S_{dqn})/|S_{dqn}|$.

\begin{figure}
    \begin{minipage}[t]{\linewidth}
    \centering
    \includegraphics[width=\linewidth]{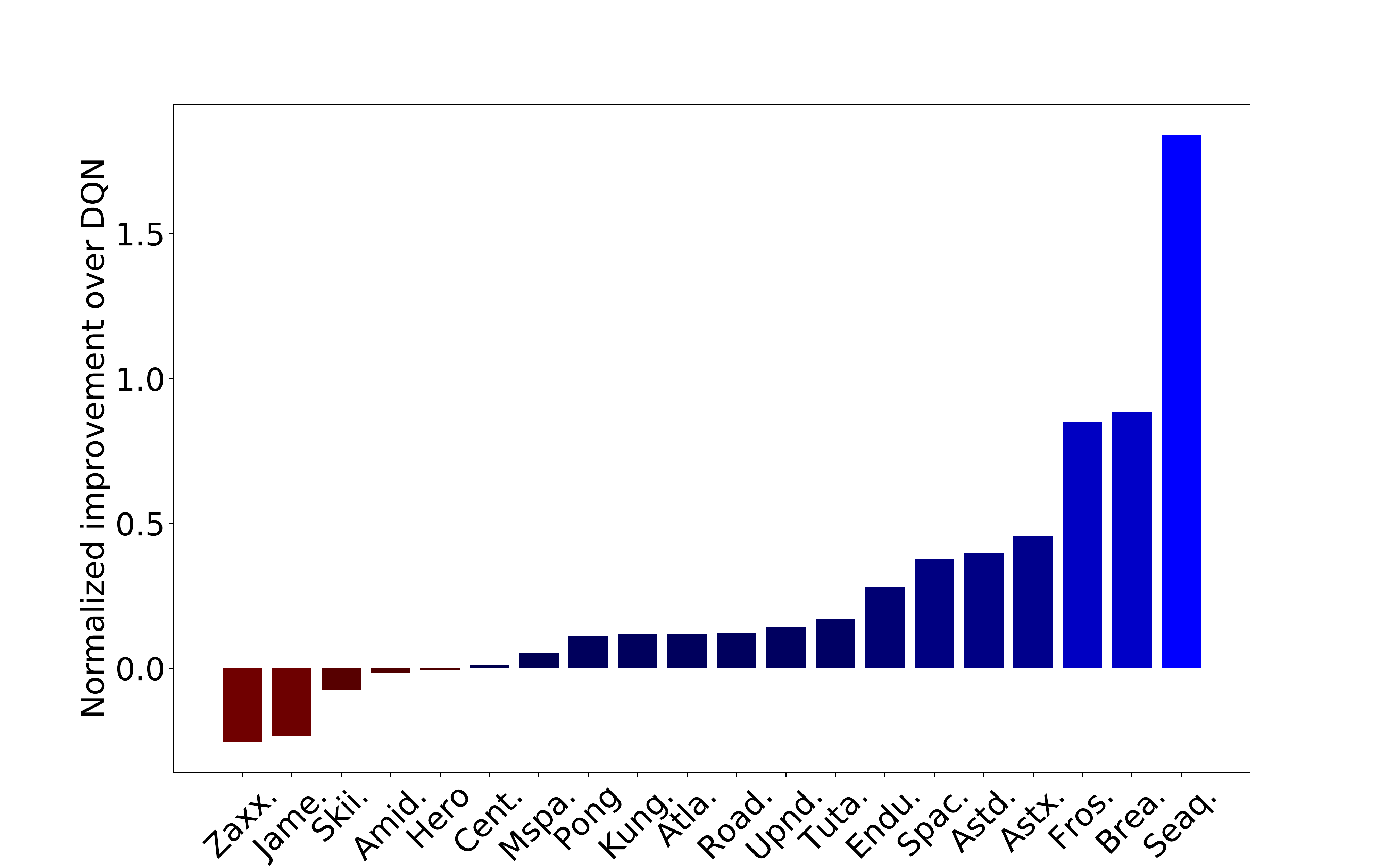}
    \caption{Normalized AUC improvement of DCPI over DQN on a subset of Atari games.}
    \label{fig:imp}
    \end{minipage}
\end{figure}

\section{Related work and discussion}

The proposed approach is related to actor-critics in general, being itself an actor-critic. It is notably related to TRPO~\cite{schulman2015trust}, that introduced a KL penalty on the greedy step as an alternative to the stochastic mixture of CPI. This is indeed very useful for continuous actions, but probably unnecessary for discrete actions, the case considered here. Moreover, TRPO is an on-policy algorithm, while the proposed DCPI approach is off-policy. This explains that we do not consider it as a baseline in Section~\ref{experiments}, but it would have been probably less sample efficient. As far as we know, there is no DQN-like TRPO algorithm, thus comparing our mixture-based DQN to one that KL-regularizes greediness would have required introducing a new algorithm.

The principle of regularizing greediness in actor-critics is quite widespread, be it with a KL divergence constraint (TRPO), a clipping of policies ratio (PPO,~\citet{schulman2017proximal}), entropy regularization (SAC), or even following policy gradient, for example. The common point of these approaches is that they focus on continuous action spaces. In the discrete case, considering a stochastic mixture is quite natural, acknowledging that its extension to the continuous case is not easy.

Performance-wise, the experiments on Cartpole show a clear improvement for DCPI over DQN: DCPI is able to reach a higher score in average, with a lower variance and a lower sensitivity to the random seed. These experiments validate the stabilizing power of CPI and its expected behaviour with respect to the mixture rate, and the consistency of the considered adaptive rates. On Atari, even if results are game-dependent, we observe an improvement on the majority of the games. Note that the improvement in score is quite clear (the score is more than doubled on some games, like Seaquest or Asterix), but the learning is not stabilized as it is in CartPole. As mentioned in Section~\ref{atari}, using a smaller (constant) mixture rate could stabilize learning and in the end increase performance, at a cost in terms of sample efficiency. This would be a problem for a single-threaded agent, like DQN, but it could improve the results of a multi-threaded agent, like R2D2~\cite{kapturowski2018recurrent}. We also recall that default used hyperparameters where better tuned for Atari than for Cartpole, and that this might also influence our empirical results. DCPI could be more efficient by better tuning its own parameters.

\section{Conclusion}

We introduced a new deep RL algorithm derived from CPI, DCPI, and this way gave a general method to regularize any pure-critic algorithm by adding a conservative actor to it, based on an approximate stochastic mixture. We gave in Section~\ref{relax} a detailed depiction of the different approximation steps we used, resulting in the end in a practical algorithm, that we evaluated on several benchmarks. We also proposed different ways to compute adaptive mixture rates for DCPI by approximating optimal rates from the literature.
Our experimental results shown, on Cartpole and on most considered Atari games, that DCPI can indeed improve the performance and the stability of learning, often at the cost of slower learning, introducing a speed/stability dilemma. We plan to investigate more adaptive rates, in order to get an even better trade-off and to be less sensitive to the new hyperparameter, and to combine the proposed approach with other variations of DQN, notably based on distributional RL, such as C51~\cite{bellemare2017distributional} or IQN~\cite{dabney2018implicit}.

\FloatBarrier

\clearpage

\bibliographystyle{aaai}
\bibliography{biblio}

\begin{thebibliography}{}

\bibitem[\protect\citeauthoryear{Barto, Sutton, and
  Anderson}{1983}]{barto1983neuronlike}
Barto, A.~G.; Sutton, R.~S.; and Anderson, C.~W.
\newblock 1983.
\newblock Neuronlike adaptive elements that can solve difficult learning
  control problems.
\newblock {\em IEEE transactions on systems, man, and cybernetics}  834--846.

\bibitem[\protect\citeauthoryear{Bellemare \bgroup et al\mbox.\egroup
  }{2013}]{bellemare2013arcade}
Bellemare, M.~G.; Naddaf, Y.; Veness, J.; and Bowling, M.
\newblock 2013.
\newblock The arcade learning environment: An evaluation platform for general
  agents.
\newblock {\em Journal of Artificial Intelligence Research} 47:253--279.

\bibitem[\protect\citeauthoryear{Bellemare, Dabney, and
  Munos}{2017}]{bellemare2017distributional}
Bellemare, M.~G.; Dabney, W.; and Munos, R.
\newblock 2017.
\newblock A distributional perspective on reinforcement learning.
\newblock In {\em Proceedings of the International Conference on Machine
  Learning},  449--458.
\newblock JMLR. org.

\bibitem[\protect\citeauthoryear{Bertsekas and
  Tsitsiklis}{1996}]{bertsekas1996neuro}
Bertsekas, D.~P., and Tsitsiklis, J.~N.
\newblock 1996.
\newblock {\em Neuro-dynamic programming}, volume~5.
\newblock Athena Scientific Belmont, MA.

\bibitem[\protect\citeauthoryear{Bradtke and Barto}{1996}]{bradtke1996linear}
Bradtke, S.~J., and Barto, A.~G.
\newblock 1996.
\newblock Linear least-squares algorithms for temporal difference learning.
\newblock {\em Machine learning} 22(1-3):33--57.

\bibitem[\protect\citeauthoryear{Brockman \bgroup et al\mbox.\egroup
  }{2016}]{brockman2016openai}
Brockman, G.; Cheung, V.; Pettersson, L.; Schneider, J.; Schulman, J.; Tang,
  J.; and Zaremba, W.
\newblock 2016.
\newblock Openai gym, 2016.
\newblock {\em arXiv preprint arXiv:1606.01540}.

\bibitem[\protect\citeauthoryear{Castro \bgroup et al\mbox.\egroup
  }{2018}]{castro2018dopamine}
Castro, P.~S.; Moitra, S.; Gelada, C.; Kumar, S.; and Bellemare, M.~G.
\newblock 2018.
\newblock Dopamine: A research framework for deep reinforcement learning.
\newblock {\em arXiv preprint arXiv:1812.06110}.

\bibitem[\protect\citeauthoryear{Dabney \bgroup et al\mbox.\egroup
  }{2018}]{dabney2018implicit}
Dabney, W.; Ostrovski, G.; Silver, D.; and Munos, R.
\newblock 2018.
\newblock Implicit quantile networks for distributional reinforcement learning.
\newblock In {\em Proceedings of the 35th International Conference on Machine
  Learning}, volume~80,  1096--1105.
\newblock PMLR.

\bibitem[\protect\citeauthoryear{Haarnoja \bgroup et al\mbox.\egroup
  }{2018}]{haarnoja2018soft}
Haarnoja, T.; Zhou, A.; Abbeel, P.; and Levine, S.
\newblock 2018.
\newblock Soft actor-critic: Off-policy maximum entropy deep reinforcement
  learning with a stochastic actor.
\newblock In {\em Proceedings of the International Conference on Machine
  Learning},  1861--1870.

\bibitem[\protect\citeauthoryear{Hessel \bgroup et al\mbox.\egroup
  }{2018}]{hessel2018rainbow}
Hessel, M.; Modayil, J.; Van~Hasselt, H.; Schaul, T.; Ostrovski, G.; Dabney,
  W.; Horgan, D.; Piot, B.; Azar, M.; and Silver, D.
\newblock 2018.
\newblock Rainbow: Combining improvements in deep reinforcement learning.
\newblock In {\em Proceedings of the 32nd AAAI Conference on Artificial
  Intelligence}.

\bibitem[\protect\citeauthoryear{Kakade and
  Langford}{2002}]{kakade2002approximately}
Kakade, S., and Langford, J.
\newblock 2002.
\newblock Approximately optimal approximate reinforcement learning.
\newblock In {\em Proceedings of the 19th International Conference on Machine
  Learning}, volume~2,  267--274.

\bibitem[\protect\citeauthoryear{Kapturowski \bgroup et al\mbox.\egroup
  }{2018}]{kapturowski2018recurrent}
Kapturowski, S.; Ostrovski, G.; Quan, J.; Munos, R.; and Dabney, W.
\newblock 2018.
\newblock Recurrent experience replay in distributed reinforcement learning.
\newblock In {\em Proceedings of the International Conference on Learning
  Representation}.

\bibitem[\protect\citeauthoryear{Kingma and Ba}{2015}]{kingma2014adam}
Kingma, D.~P., and Ba, J.
\newblock 2015.
\newblock Adam: A method for stochastic optimization.
\newblock In {\em Proceedings of the 3rd International Conference for Learning
  Representations}.

\bibitem[\protect\citeauthoryear{Machado \bgroup et al\mbox.\egroup
  }{2018}]{machado2018revisiting}
Machado, M.~C.; Bellemare, M.~G.; Talvitie, E.; Veness, J.; Hausknecht, M.; and
  Bowling, M.
\newblock 2018.
\newblock Revisiting the arcade learning environment: Evaluation protocols and
  open problems for general agents.
\newblock {\em Journal of Artificial Intelligence Research} 61:523--562.

\bibitem[\protect\citeauthoryear{Mnih \bgroup et al\mbox.\egroup
  }{2015}]{mnih2015human}
Mnih, V.; Kavukcuoglu, K.; Silver, D.; Rusu, A.~A.; Veness, J.; Bellemare,
  M.~G.; Graves, A.; Riedmiller, M.; Fidjeland, A.~K.; Ostrovski, G.; et~al.
\newblock 2015.
\newblock Human-level control through deep reinforcement learning.
\newblock {\em Nature} 518(7540):529.

\bibitem[\protect\citeauthoryear{Ostrovski \bgroup et al\mbox.\egroup
  }{2017}]{ostrovski2017count}
Ostrovski, G.; Bellemare, M.~G.; van~den Oord, A.; and Munos, R.
\newblock 2017.
\newblock Count-based exploration with neural density models.
\newblock In {\em Proceedings of the 34th International Conference on Machine
  Learning},  2721--2730.

\bibitem[\protect\citeauthoryear{Pirotta \bgroup et al\mbox.\egroup
  }{2013}]{pirotta2013safe}
Pirotta, M.; Restelli, M.; Pecorino, A.; and Calandriello, D.
\newblock 2013.
\newblock Safe policy iteration.
\newblock In {\em Proceedings of the 30th International Conference on Machine
  Learning},  307--315.

\bibitem[\protect\citeauthoryear{Puterman and
  Shin}{1978}]{puterman1978modified}
Puterman, M.~L., and Shin, M.~C.
\newblock 1978.
\newblock Modified policy iteration algorithms for discounted markov decision
  problems.
\newblock {\em Management Science} 24(11):1127--1137.

\bibitem[\protect\citeauthoryear{Puterman}{1994}]{puterman1994markov}
Puterman, M.~L.
\newblock 1994.
\newblock {\em Markov decision processes: discrete stochastic dynamic
  programming}.
\newblock John Wiley \& Sons.

\bibitem[\protect\citeauthoryear{Scherrer and Geist}{2014}]{scherrer2014local}
Scherrer, B., and Geist, M.
\newblock 2014.
\newblock Local policy search in a convex space and conservative policy
  iteration as boosted policy search.
\newblock In {\em Proceedings of the Joint European Conference on Machine
  Learning and Knowledge Discovery in Databases},  35--50.
\newblock Springer.

\bibitem[\protect\citeauthoryear{Scherrer \bgroup et al\mbox.\egroup
  }{2015}]{scherrer2015approximate}
Scherrer, B.; Ghavamzadeh, M.; Gabillon, V.; Lesner, B.; and Geist, M.
\newblock 2015.
\newblock Approximate modified policy iteration and its application to the game
  of tetris.
\newblock {\em Journal of Machine Learning Research} 16:1629--1676.

\bibitem[\protect\citeauthoryear{Scherrer}{2014}]{scherrer2014approximate}
Scherrer, B.
\newblock 2014.
\newblock Approximate policy iteration schemes: a comparison.
\newblock In {\em Proceedings of the 31st International Conference on Machine
  Learning},  1314--1322.

\bibitem[\protect\citeauthoryear{Schulman \bgroup et al\mbox.\egroup
  }{2015}]{schulman2015trust}
Schulman, J.; Levine, S.; Abbeel, P.; Jordan, M.; and Moritz, P.
\newblock 2015.
\newblock Trust region policy optimization.
\newblock In {\em Proceedings of the 32nd International Conference on Machine
  Learning},  1889--1897.

\bibitem[\protect\citeauthoryear{Schulman \bgroup et al\mbox.\egroup
  }{2017}]{schulman2017proximal}
Schulman, J.; Wolski, F.; Dhariwal, P.; Radford, A.; and Klimov, O.
\newblock 2017.
\newblock Proximal policy optimization algorithms.
\newblock {\em arXiv preprint arXiv:1707.06347}.

\bibitem[\protect\citeauthoryear{Sutton}{1988}]{sutton1988learning}
Sutton, R.~S.
\newblock 1988.
\newblock Learning to predict by the methods of temporal differences.
\newblock {\em Machine learning} 3(1):9--44.

\end{thebibliography}

\iftrue

\FloatBarrier

\newpage
\appendix
\onecolumn

\section{Theoretical insights \label{app:analysis}}

Relaxing conservative policy iteration leads to a quite different algorithmic scheme, even if the essence of CPI is kept. Here, we provide a (partial and preliminary) analysis of the propagation of errors in scheme~\eqref{cpi_4}, written here with value functions for the sake of analysis (that does not change fundamentally things),
\begin{equation}
\begin{cases}
    \pi_{k+1} = (1 - \alpha_{k+1})\pi_k  + \alpha_{k+1}\pi'_{k+1} + \p{\epsilon_{k+1}}
    \\
    v_{k+1} = T_{\pi_{k+1}}^m v_{k} + \epsilon_{k+1}
\end{cases}, \text{with }  \pi'_{k+1} \in \gr v_k.
\end{equation}
The analysis we propose mimics the one of approximate modified policy iteration (AMPI)~\cite{scherrer2015approximate}, and borrows the terms introduced there. What we do here is generalizing (up to the different error in policy) their Lemma~2. Then, we'll discuss what may be the consequences of this result, without doing the full propagation analysis, which is quite tedious.

The goal is to bound the \emph{loss} $l_k = v_* - v_{\pi_k} \geq 0$. It can be decomposed as follows,
\begin{equation}
    l_k =  v_* - v_{\pi_k} = v_* - T_{\pi_k}^m v_{k-1} + T_{\pi_k}^m v_{k-1} - v_{\pi_k}.
\end{equation}
We introduce the \emph{distance} $d_k = v_* - T_{\pi_k}^m v_{k-1} = v_* - (v_k - \epsilon_k)$ and the \emph{shift} $s_k = T_{\pi_k}^m v_{k-1} - v_{\pi_k} = (v_k - \epsilon_k)  - v_{\pi_k}$. We also introduce the \emph{Bellman residual} $b_k = v_k - T_{\pi_{k+1}} v_k$. With this notations, the loss rewrites $l_k = d_k + s_k$.
The core of the analysis of AMPI consists in proving point-wise inequalities for $b_k$, $s_k$ and $d_k$~\cite[Lemma~ 2]{scherrer2015approximate}. We provide a similar result here.

\begin{theorem}
\label{th:bound}
Let $k\geq 1$ and define 
\begin{align}
    x_k &= (I-\gamma P_{\pi_k}) \epsilon_k - \langle q_k, \epsilon'_{k+1}\rangle \text{ and}
    \\
    y_k &= -\left((1-\alpha_k) \gamma P_{\pi_k} + \alpha_k \gamma P_{\pi_*}\right)\epsilon_k - \langle q_k, \epsilon'_{k+1}\rangle,
    \\
    \text{with }
    \langle q_k, \epsilon'_{k+1}\rangle &= \left(\sum_a \epsilon'_{k+1}(s,a)\left(r(s,a)+\gamma\mathbb{E}_{s'|s,a} [v_k(s')]\right)\right)_{s\in\mathcal{S}}.
\end{align}
We have:
\begin{align}
    b_k &\leq (\gamma P_{\pi_k})^m b_{k-1} + x_k,
    \\
    d_{k+1} &\leq \left((1-\alpha_{k+1}) I + \alpha_{k+1} \gamma P_{\pi_*}\right) d_k + y_k + \sum_{j=1}^{m-1} (\gamma P_{\pi_{k+1}})^j b_k + (1-\alpha_{k+1}) (\gamma P_{\pi_k})^m b_{k-1},
    \\
    s_k &= (\gamma P_{\pi_k})^m (I-\gamma P_{\pi_k})^{-1} b_{k-1}.
\end{align}
\end{theorem}
\begin{proof}
    The bound for $s_k$ is obtained as for AMPI, we give it for completeness:
    \begin{align}
        s_k &= T_{\pi_k}^m v_{k-1} - v_{\pi_k}
        \\
        &= T_{\pi_k}^m v_{k-1} - T_{\pi_k}^{m+\infty} v_{k-1}
        \\
        &= (\gamma P_{\pi_k})^m (v_{k-1} - T_{\pi_k}^\infty v_{k-1})
        \\
        &= (\gamma P_{\pi_k})^m \sum_{j=0}^\infty (T_{\pi_k}^{j} v_{k-1} - T_{\pi_k}^{j+1} v_{k-1})
        \\
        &= (\gamma P_{\pi_k})^m \sum_{j=0}^\infty  (\gamma P_{\pi_k})^j (v_{k-1} - T_{\pi_k} v_{k-1})
        \\
        &= (\gamma P_{\pi_k})^m (I-\gamma P_{\pi_k})^{-1} b_{k-1}.
    \end{align}
    The bounds for $b_k$ and $d_k$ are a bit different, but rely on the same decomposition principle (taking into account that the new policy is a mixture, here). We start by bounding $b_k$:
    \begin{align}
        b_k &= v_k - T_{\pi_{k+1}} v_k
        \\
        &= v_k - T_{\pi_k} v_k + T_{\pi_k} v_k - \left( (1-\alpha_{k+1}) T_{\pi_k} v_k + \alpha_{k+1} T_{\pi'_{k+1}} v_k + \langle q_k, \epsilon'_{k+1}\rangle \right)
        \\
        &= v_k - T_{\pi_k} v_k + \alpha_{k+1} \underbrace{\left(T_{\pi_k} v_k - T_{\pi'_{k+1}} v_k\right)}_{\leq 0} - \langle q_k, \epsilon'_{k+1}\rangle
        \\
        &\leq v_k - T_{\pi_k} v_k - \langle q_k, \epsilon'_{k+1}\rangle
        \\
        &= \underbrace{v_k - \epsilon_k}_{=T_{\pi_k}^m v_{k-1}} + \epsilon_k - T_{\pi_k}(v_k - \epsilon_k) - \gamma P_{\pi_k}\epsilon_k - \langle q_k, \epsilon'_{k+1}\rangle
        \\
        &= \underbrace{T_{\pi_k}^m v_{k-1} - T_{\pi_k}^{m+1} v_{k-1}}_{=(\gamma P_{\pi_k})^m (v_{k-1} - T_{\pi_k}^{m} v_{k-1})} + \underbrace{(I-\gamma P_{\pi_k}) \epsilon_k - \langle q_k, \epsilon'_{k+1}\rangle}_{=x_k}
        \\
        &= (\gamma P_{\pi_k})^m b_{k-1} + x_k.
    \end{align}
    The bound on $d_k$ requires the following equalities:
    \begin{align}
        T_{\pi_k}^m v_{k-1} - T_{\pi_k} v_k &= T_{\pi_k}^m v_{k-1} - T_{\pi_k} (T_{\pi_k}^m v_{k-1} + \epsilon_k)
        \\
        &= T_{\pi_k}^m v_{k-1} - T_{\pi_k}^{m+1} v_{k-1} - \gamma P_{\pi_k} \epsilon_k
        \\
        &= (\gamma P_{\pi_k})^m b_{k-1} - \gamma P_{\pi_k}\epsilon_k,
    \end{align}
    and
    \begin{align}
        T_{\pi_{k+1}} v_{k} - T_{\pi_{k+1}}^m v_{k} &= \sum_{j=1}^{m-1} (T_{\pi_{k+1}}^j v_{k} -  T_{\pi_{k+1}}^{j+1} v_{k}) 
        \\
        &= \sum_{j=1}^{m-1} (\gamma P_{\pi_{k+1}})^j (v_k - T_{\pi_{k+1}} v_k
        \\
        &= \sum_{j=1}^{m-1} (\gamma P_{\pi_{k+1}})^j b_k.
    \end{align}
    We can now bound $d_k$:
    \begin{align}
        d_{k+1} &= v_* - T_{\pi_{k+1}}^m v_k
        \\
        &= v_* - T_{\pi_{k+1}} v_k + T_{\pi_{k+1}} v_k - T_{\pi_{k+1}}^m v_k
        \\
        &= v_* - \left( (1-\alpha_{k+1}) T_{\pi_k} v_k + \alpha_{k+1} T_{\pi'_{k+1}} v_k + \langle q_k, \epsilon'_{k+1}\rangle \right) + \sum_{j=1}^{m-1} (\gamma P_{\pi_{k+1}})^j b_k
        \\
        &= (1-\alpha_{k+1})(v_* -  T_{\pi_k} v_k) + \alpha_{k+1} (v_* - T_{\pi'_{k+1}} v_k) - \langle q_k, \epsilon'_{k+1}\rangle + \sum_{j=1}^{m-1} (\gamma P_{\pi_{k+1}})^j b_k.
    \end{align}
    We have that
    \begin{align}
        v_* -  T_{\pi_k} v_k &= 
        \underbrace{v_* - T_{\pi_k}^m v_{k-1}}_{=d_k} + 
        \underbrace{T_{\pi_k}^m v_{k-1} - T_{\pi_k} v_k}_{=(\gamma P_{\pi_k})^m b_{k-1} - \gamma P_{\pi_k}\epsilon_k}
        \\ \text{and that }
        v_* - T_{\pi'_{k+1}} v_k &=
        \underbrace{v_* - T_{\pi_*} v_k }_{= \gamma P_{\pi_*}(v_* - v_k) = \gamma P_{\pi_*}(d_k - \epsilon_k)}  + 
        \underbrace{T_{\pi_*} v_k - T_{\pi'_{k+1}} v_k}_{\leq 0}.
    \end{align}
    Injecting this in the preceding equation, rearranging terms and writing $y_k = -((1-\alpha_k) \gamma P_{\pi_k} + \alpha_k \gamma P_{\pi_*}\epsilon_k) - \langle q_k, \epsilon'_{k+1}\rangle$, we obtain the stated result,
    \begin{equation}
        d_{k+1} \leq \left((1-\alpha_{k+1}) I + \alpha_{k+1} \gamma P_{\pi_*}\right) d_k + y_k + \sum_{j=1}^{m-1} (\gamma P_{\pi_{k+1}}^j)^j b_k + (1-\alpha_{k+1}) (\gamma P_{\pi_k})^m b_{k-1}.
    \end{equation}
\end{proof}

Setting $\alpha_k=1$, up to the fact that we consider a different error in the policy approximation (that does not change how the errors would propagate), we retrieve the result of Lemma~2 in~\cite{scherrer2015approximate}. This lemma is the core building block of the analysis for propagation of errors in AMPI, so in principle we could build a bound on $\|l_k\|$ (involving concentrability coefficients, horizon and error terms), for some weighted norm, based on Thm.~\ref{th:bound}. However, it would be much more involved and tedious than for AMPI, mainly due to the term $(1-\alpha_{k+1}) I$. Therefore, we do not push the analysis further, but we discuss a bit the result it would give.

Without errors ($\epsilon_k = \epsilon'_k = 0$), Thm.~1 shows that $v_{\pi_k}$ will converge linearly to $v_*$. With $\alpha_k=1$ (this corresponds to MPI), the leading term (multiplying $d_k$) is $\gamma P_{\pi_*}$, that gives a $\gamma$-contraction and leads to a bound $\|v_* - v_{\pi_k}\|_\infty = \mathcal{O}(\gamma^k)$. With $\alpha_k<1$, the leading term is $(1-\alpha_{k+1}) I + \alpha_{k+1} \gamma P_{\pi_*}$, that gives an $\eta_k$-contraction with $\eta_k=1-\alpha_k(1-\gamma)$. If $\alpha_k$ does not goes to fast towards zero, this would also gives a linear convergence. Indeed, using the fact that $\ln(1-x) \leq -x$ for $x\in(0,1)$,
\begin{equation}
    \prod_{i=1}^k \eta_k = \exp \sum_{i=1}^k \ln \eta_i = \exp \sum_{i=1}^k \ln(1 - \alpha_i(1-\gamma)) \leq \exp(-(1-\gamma)\sum_{i=1}^k \alpha_i).
\end{equation}
Therefore, this would lead to a bound $\|v_* - v_{\pi_k}\|_\infty = \mathcal{O}(\prod_{i=1}^k \eta_i) = \mathcal{O}(\exp(-(1-\gamma)\sum_{i=1}^k \alpha_i))$. If we still have a linear convergence, it is slower as long as $\alpha_k<1$, which was to be expected without approximation error. However, at least this scheme does not break convergence.

With errors, we conjecture that we would obtain a bound close to the one of AMPI~\cite[Thm.~7]{scherrer2015approximate}, maybe with a larger propagation of errors (much like the convergence is slower, in the exact case), and so worse than the original bound of CPI~\cite{kakade2002approximately,scherrer2014approximate} (notably, with bigger concentrability coefficient). This is to be expected, the bound of CPI relies heavily on using $m=\infty$, on how the approximation error is plugged in the approximate dynamic scheme, and on using the $d_{\pi,\mu}$ distribution to sample transitions for learning approximations, three things that we relax. Yet, we still think that relaxing greediness is worth experimentally speaking, and that much remains to be done regarding its theoretical understanding.

\section{Experimental details \label{app:details}}
In this appendix we provide additional details about the experiments.%

\subsection{SPI rate on Cartpole}

Figure~\ref{fig:spi_totol} reports the result of combining DCPI with the SPI adaptive rate described in Eq.~\eqref{alpha_spi}. As this rate is not bounded by 1, we clip $\alpha_k$ between $0$ and $1$. The behavior is similar to the one of $\alpha^{adx}$.

\begin{figure}
    \centering
    \includegraphics[width=.7\linewidth]{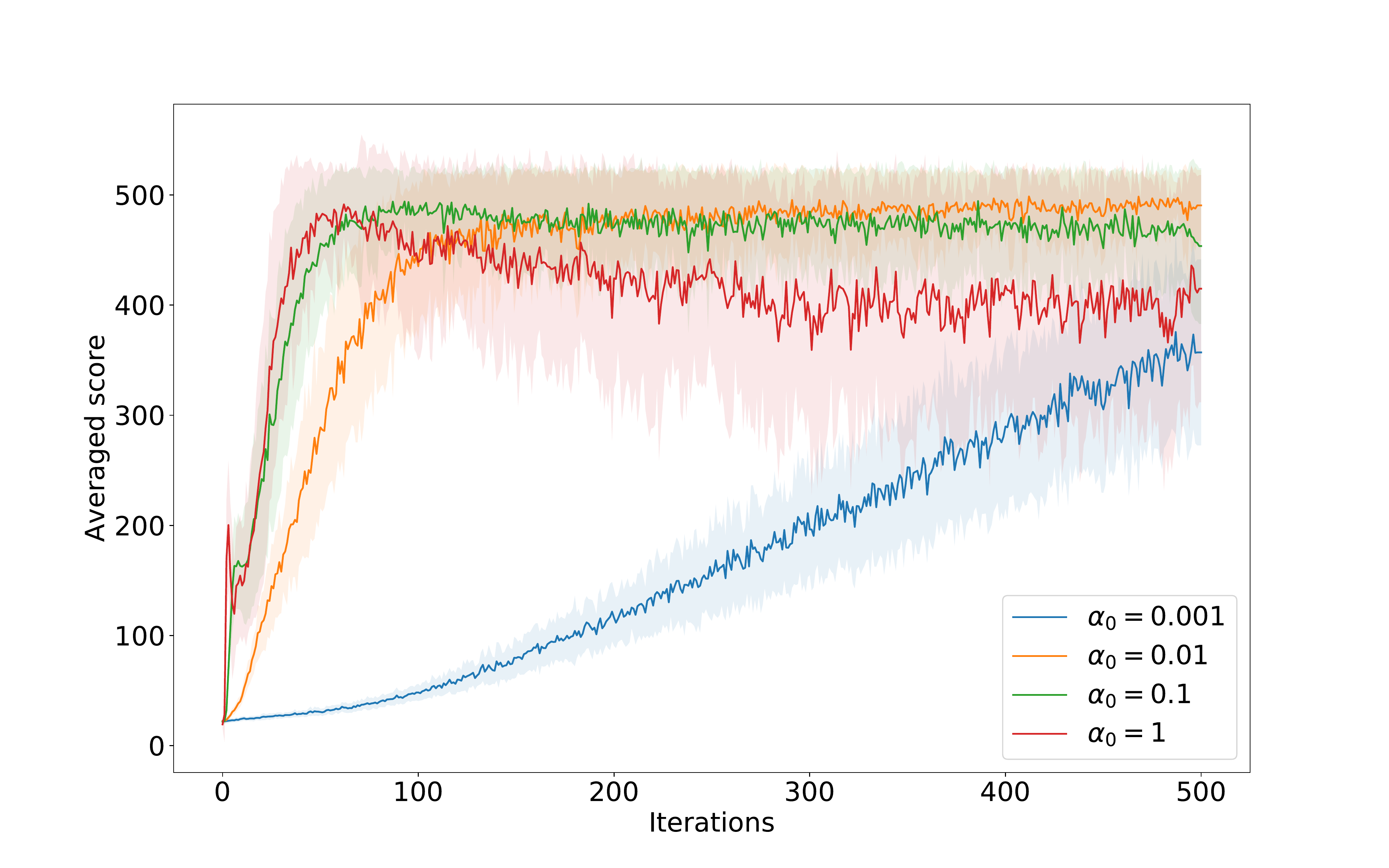}
    \caption{Training curves for DCPI on Cartpole with SPI adaptive rates and different values of $\alpha_0$. This exhibits almost the same behaviour as using its variant $\alpha^{adx}$.}
    \label{fig:spi_totol}
\end{figure}

\subsection{Parameters}

In Tables~\ref{tab:cartpole} and~\ref{tab:atari}, we give the hyperparemeters used for our experiments, including networks architecture. We use the following notations to describe neural networks: $\FC n$ is a fully connected layer with $n$ neurons; $\Conv_{a,b}^{d}c$ is a 2d convolutional layer with $c$ filters of size $a  \times b$ and a stride of~$d$.

\begin{table}%
    \centering
    \caption{Parameters used for DCPI on Cartpole.}
    \begin{tabular}{ll}
    \hline
    Parameter     & Value \\
    \hline
    $C$ (update period)    & 100\\
    $F$ (interaction  period)    & 4\\
    $\gamma$ (discount) & 0.99\\
    $|\mathcal{B}|$ & $5\cdot10^4$\\
    $|B_{\pi,k}|$ and $|B_{q,k}|$ (batch size) & 128 \\
    $\varepsilon$ (random actions rate) & 0.01\\
    $\beta_1$ & 0.99\\
    $\beta_2$ & 0.9999\\
    q-network stucture & $\FC512-\FC512-\FC8$ \\
    policy-network structure & $\FC512-\FC512-\FC8$\\
    activations & Relu\\
    optimizers & Adam ($lr=0.001$)\\
    \hline
    \end{tabular}
    \label{tab:cartpole}
\end{table}

\begin{table}%
    \centering
    \caption{Parameters used for DCPI on Atari. NB: the size of the last fully-connected layer in the q-network and policy network is the number of actions, which varies from game to game.}
    \begin{tabular}{ll}
    \hline
    Parameter     & Value \\
    \hline
    $C$ (update period)    & 8000\\
    $F$ (interaction  period)    & 4\\
    $\gamma$ (discount) & 0.99\\
    $|\mathcal{B}|$ & $10^6$\\
    $|B_{\pi,k}|$ and $|B_{q,k}|$ (batch size) & 32 \\
    $\varepsilon$ (random actions rate) & 0.01 (with a linear decay of period $2.5\cdot10^5$ steps)\\
    $\beta_1$ & 0.9999\\
    $\beta_2$ & 0.999999\\
    q-network stucture & $\Conv_{8,8}^{4}32-\Conv_{4,4}^{2}64-\Conv_{3,3}^{1}64-\FC512-\FC$\\
    policy-network structure & $\Conv_{8,8}^{4}32-\Conv_{4,4}^{2}64-\Conv_{3,3}^{1}64-\FC512-\FC$\\
    activations & Relu\\
    optimizers & RMSprop ($lr=0.00025$) \\
    \hline
    \end{tabular}
    \label{tab:atari}
\end{table}

\subsection{Full Atari}

In Figures~\ref{fig:full} and~\ref{fig:full2}, the full results of our experiment on Atari are reported. On Atari, we considered the following games, listed in three categories. Easy exploration: Asterix, Asteroids, Atlantis, Breakout, Centipede, Enduro, Jamesbond, Pong, Skiing, SpaceInvaders; Score exploit: KungFuMaster, RoadRunner, Seaquest, Tutankham, UpNDown; Hard explore (dense reward): Amidar, Frostbite, Hero, MsPacman, Zaxxon.

\begin{figure}
\begin{center}
\begin{tabular}{cc}
     \includegraphics[width=0.42\linewidth]{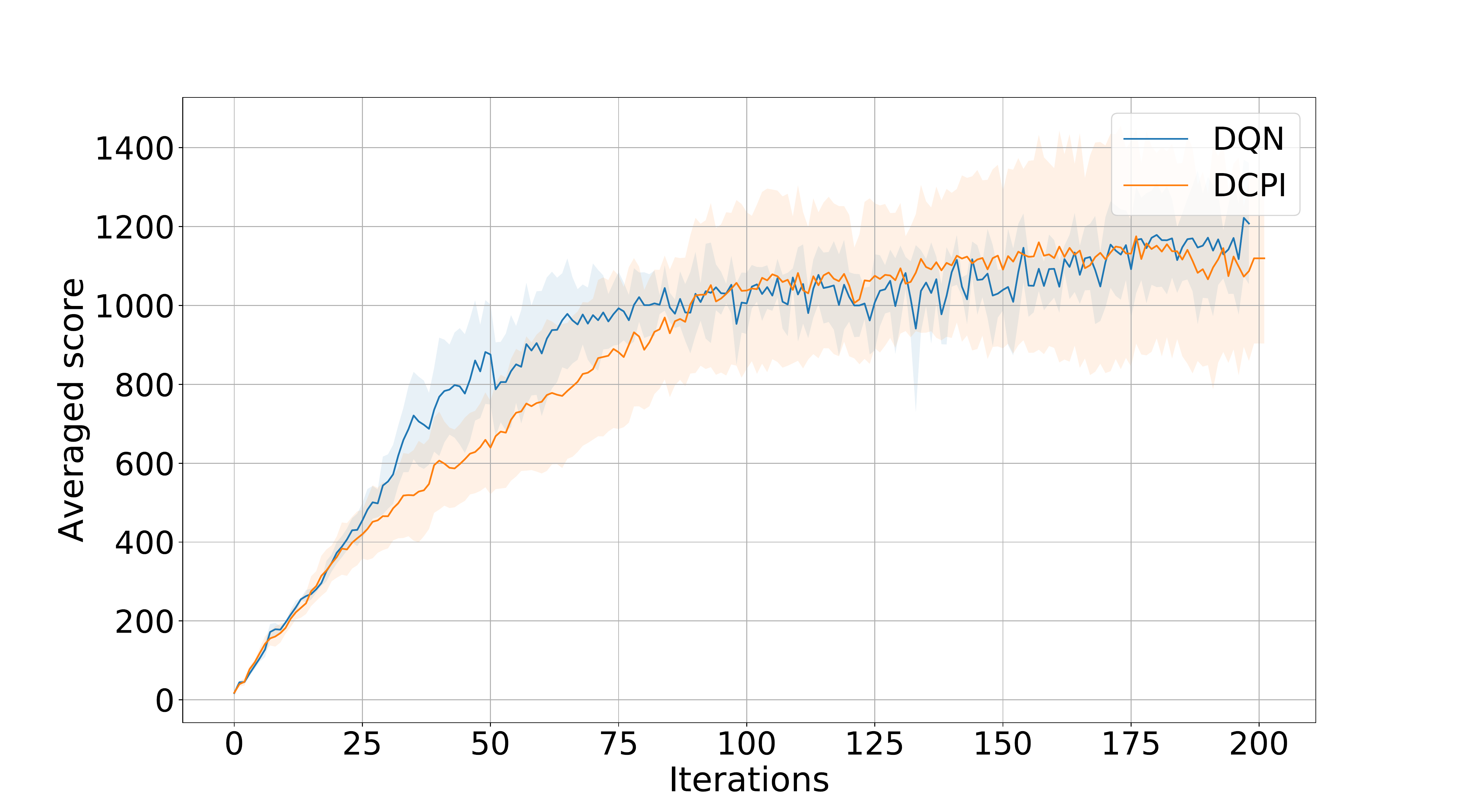} & \includegraphics[width=0.42\linewidth]{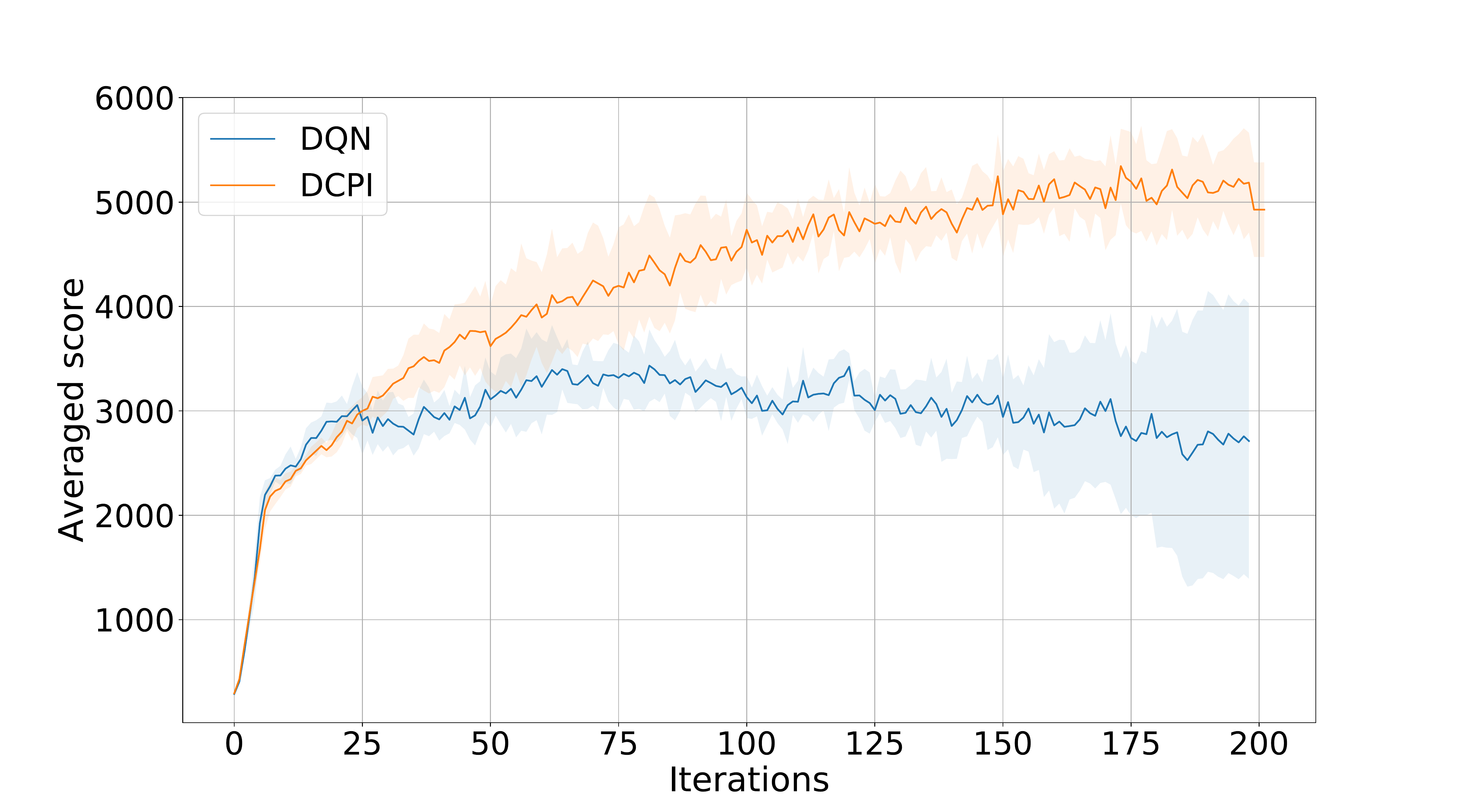}  \\
     Amidar & Asterix \\ 
     \includegraphics[width=0.42\linewidth]{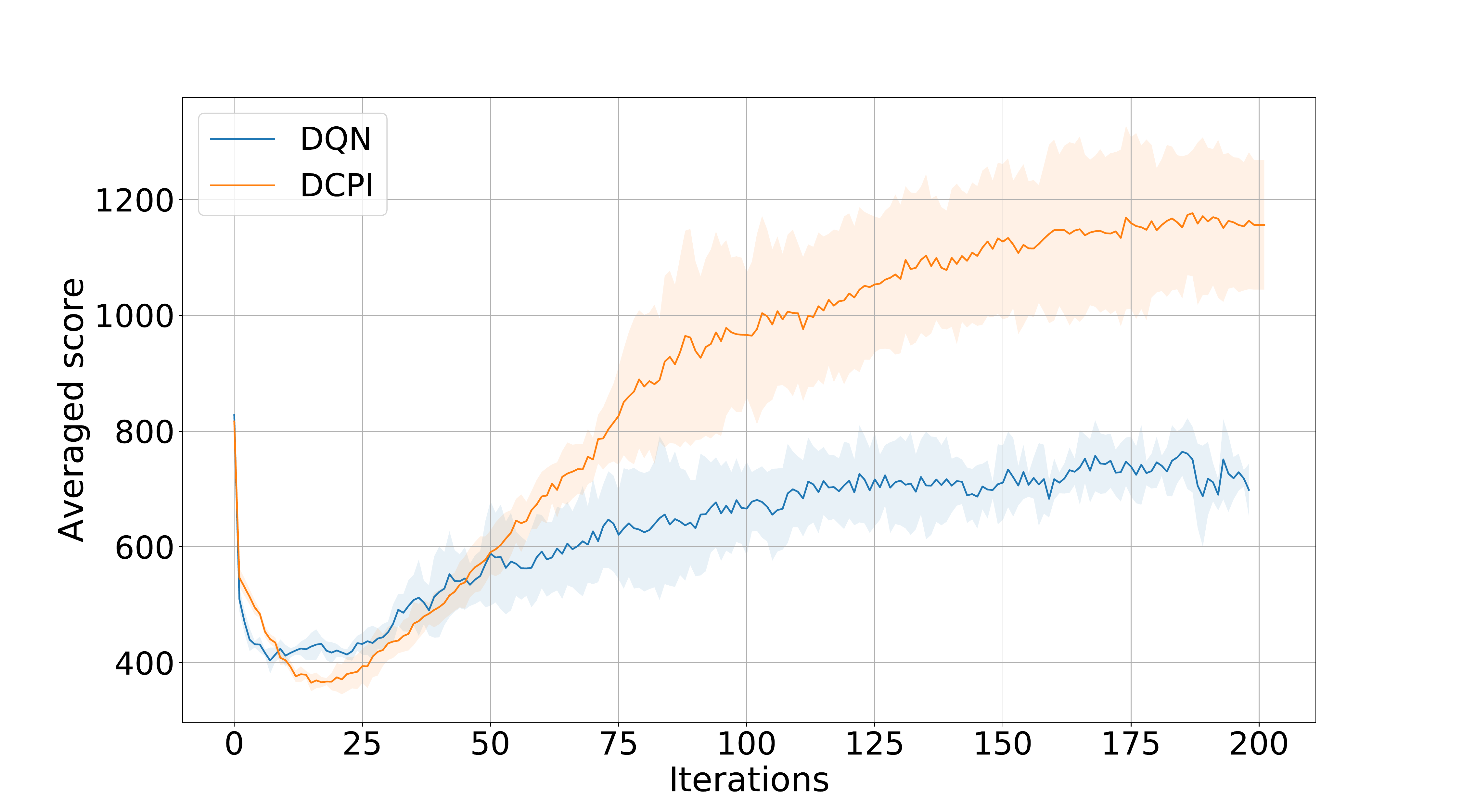} & \includegraphics[width=0.42\linewidth]{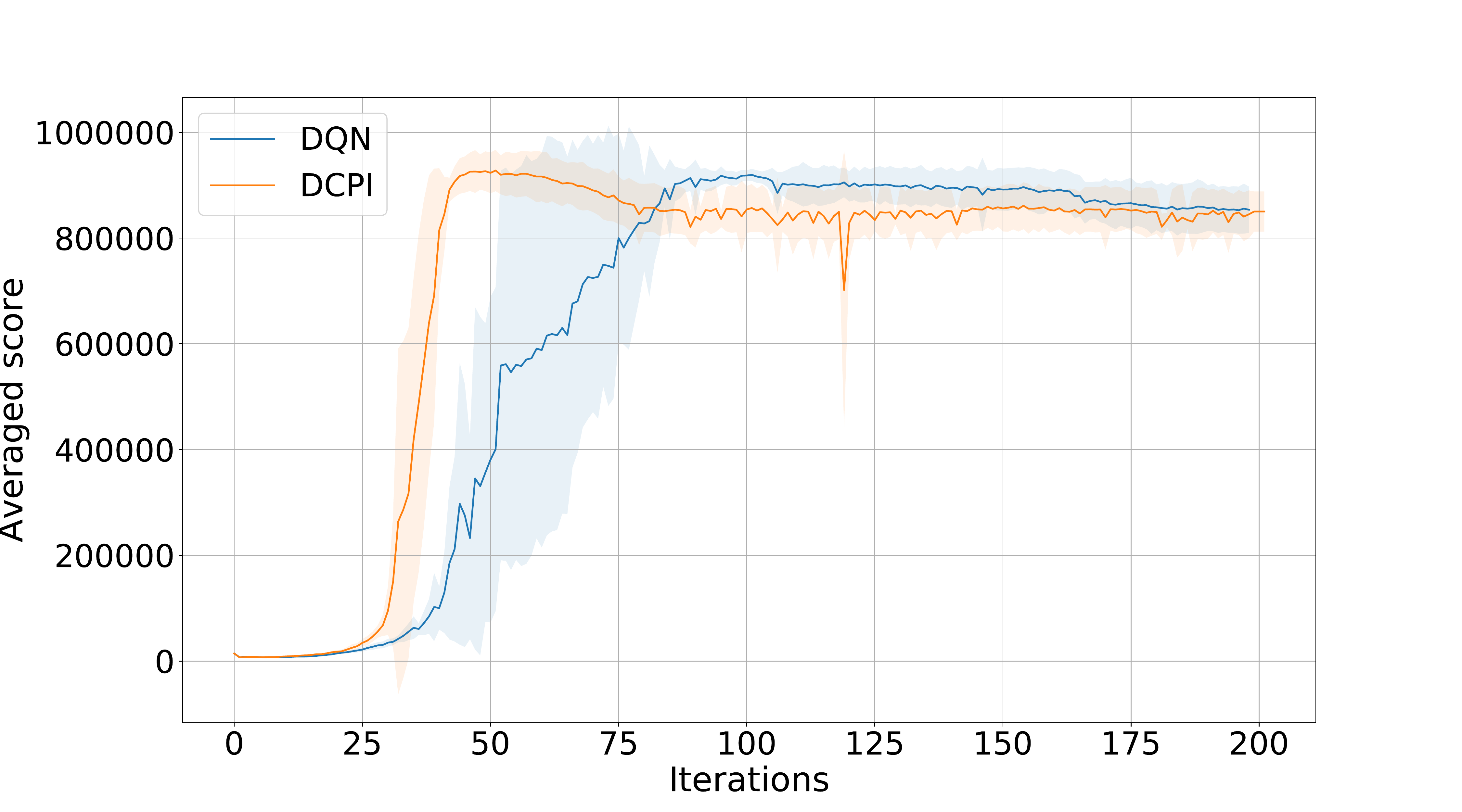}  \\
     Asteroids & Atlantis \\
     \includegraphics[width=0.42\linewidth]{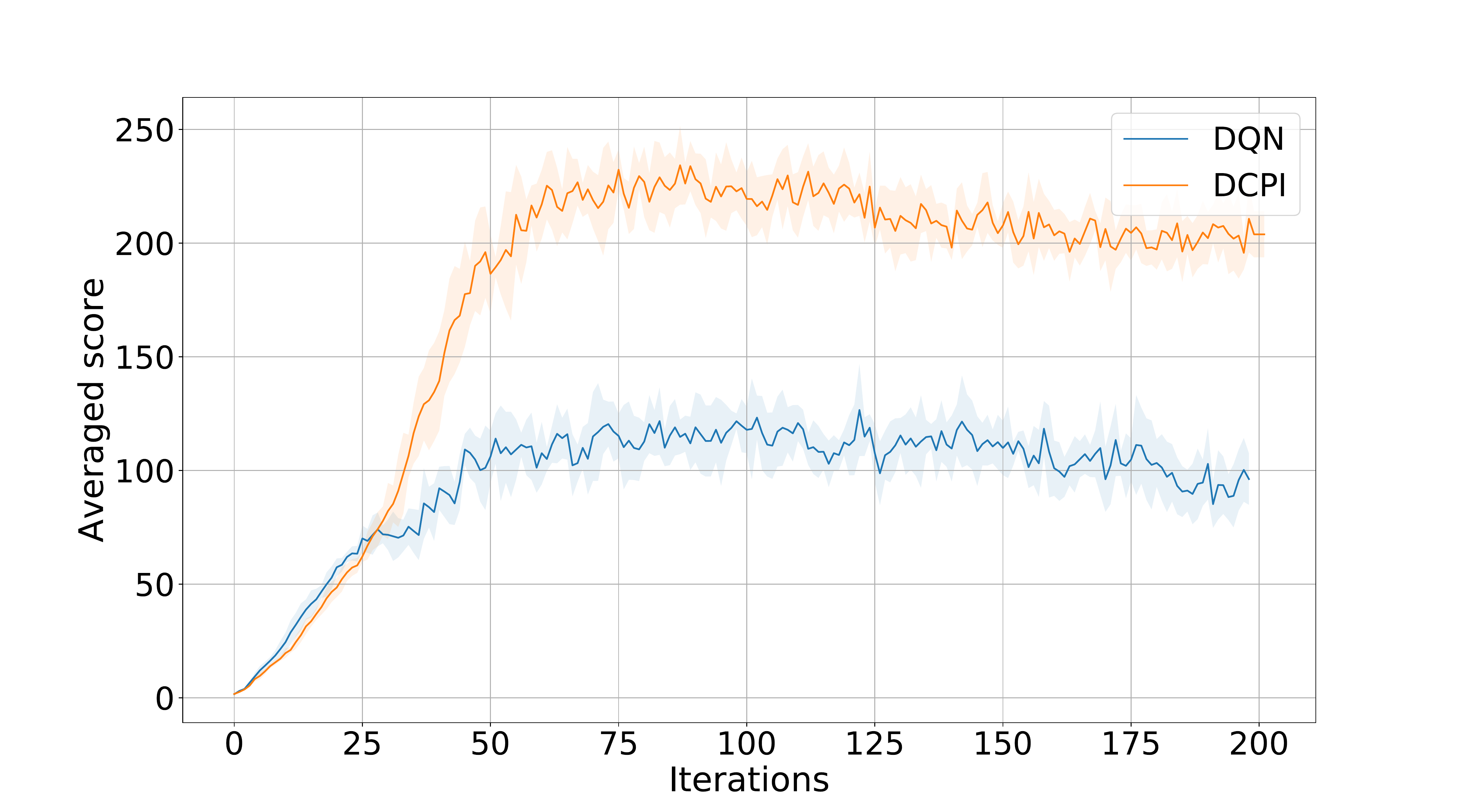} & \includegraphics[width=0.42\linewidth]{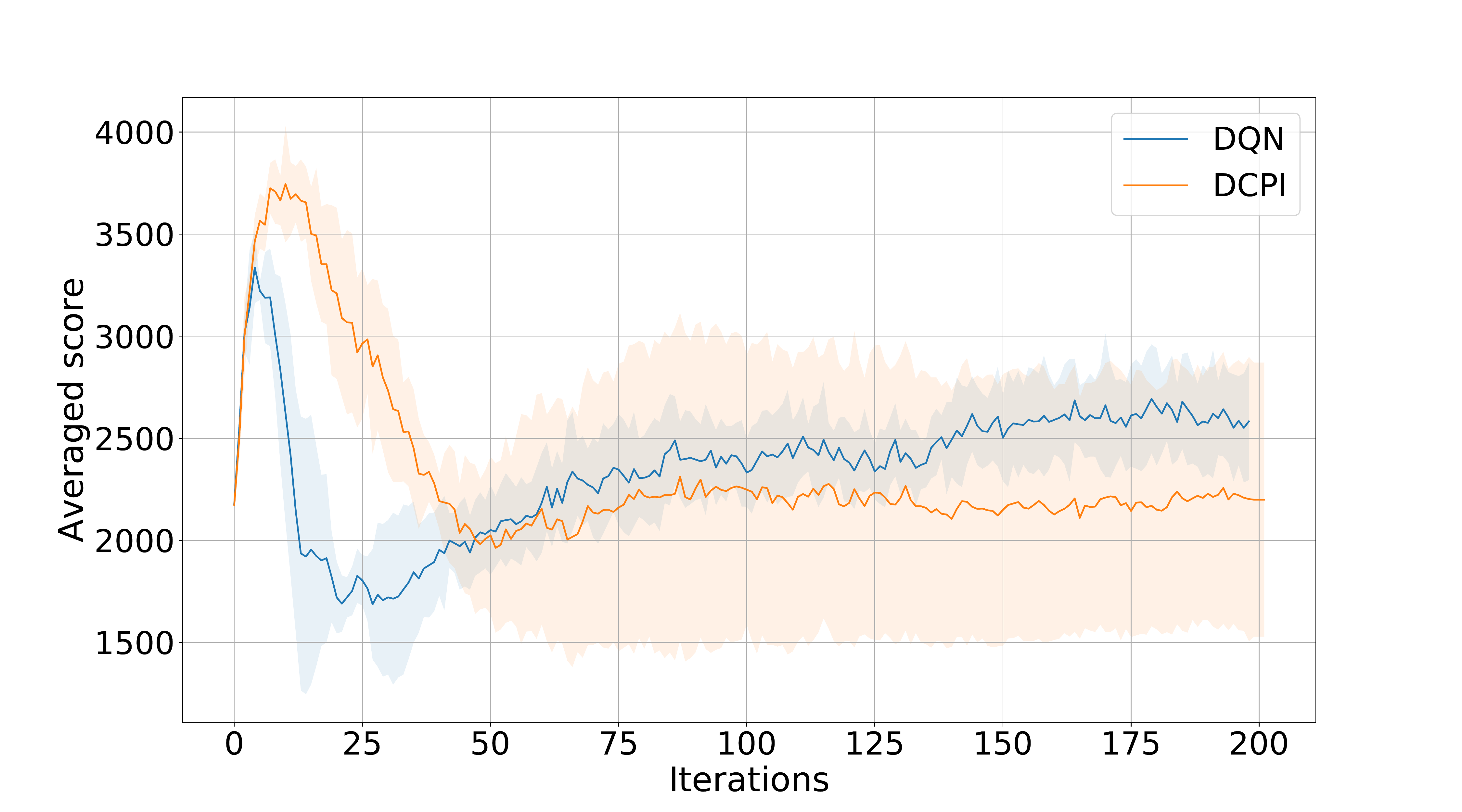}  \\
     Breakout & Centipede \\
     \includegraphics[width=0.42\linewidth]{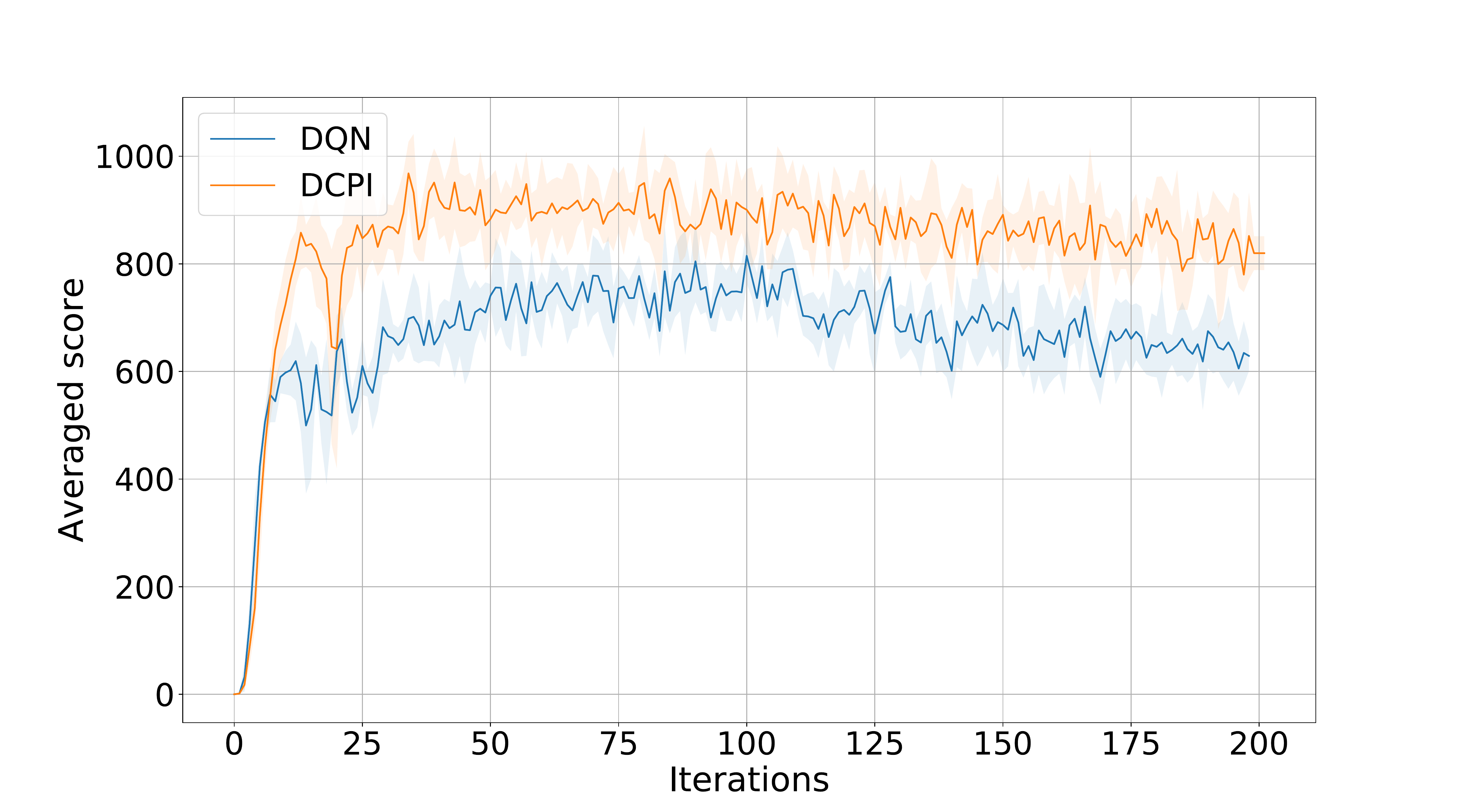} & \includegraphics[width=0.42\linewidth]{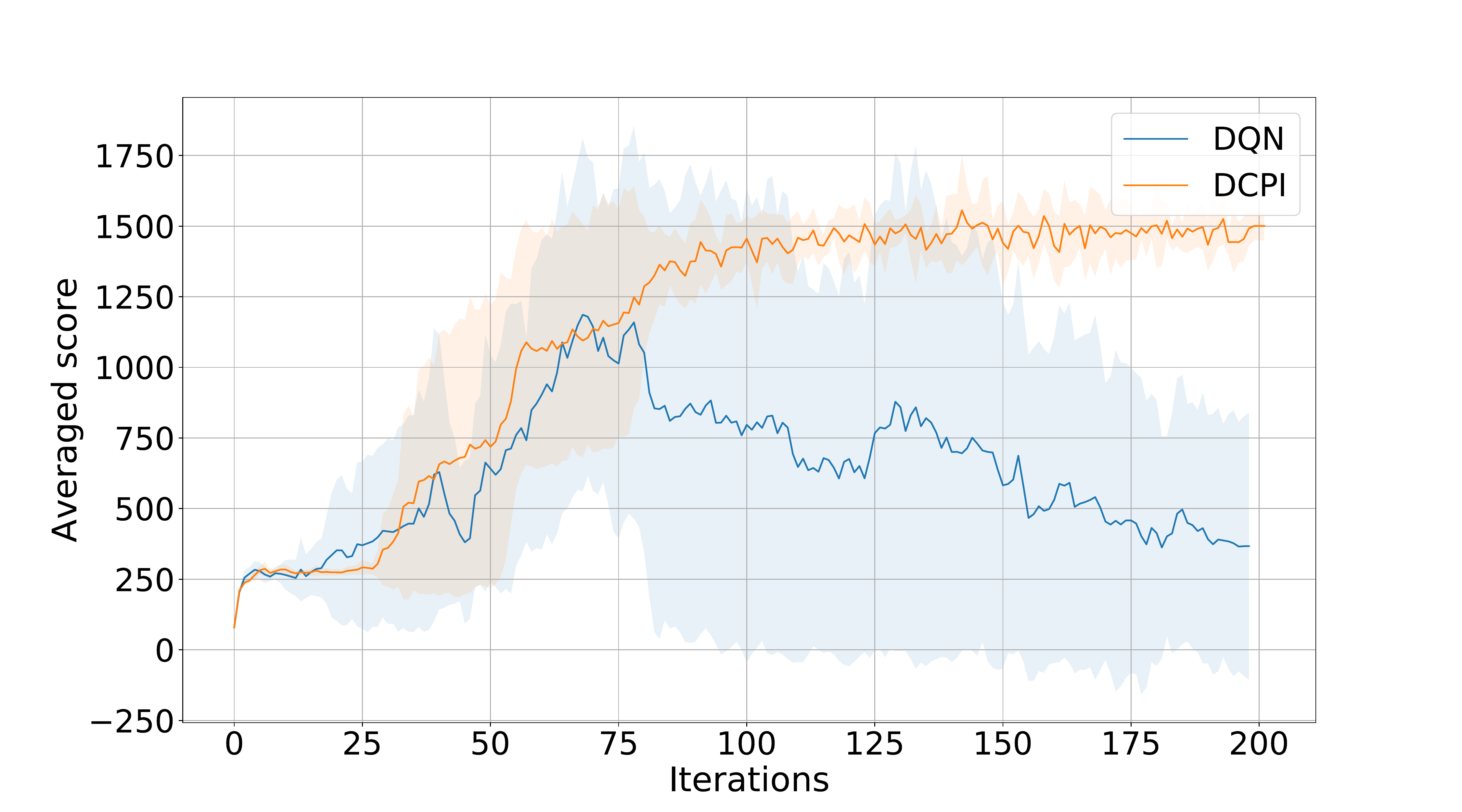}  \\
     Enduro & Frostbite \\
     \includegraphics[width=0.42\linewidth]{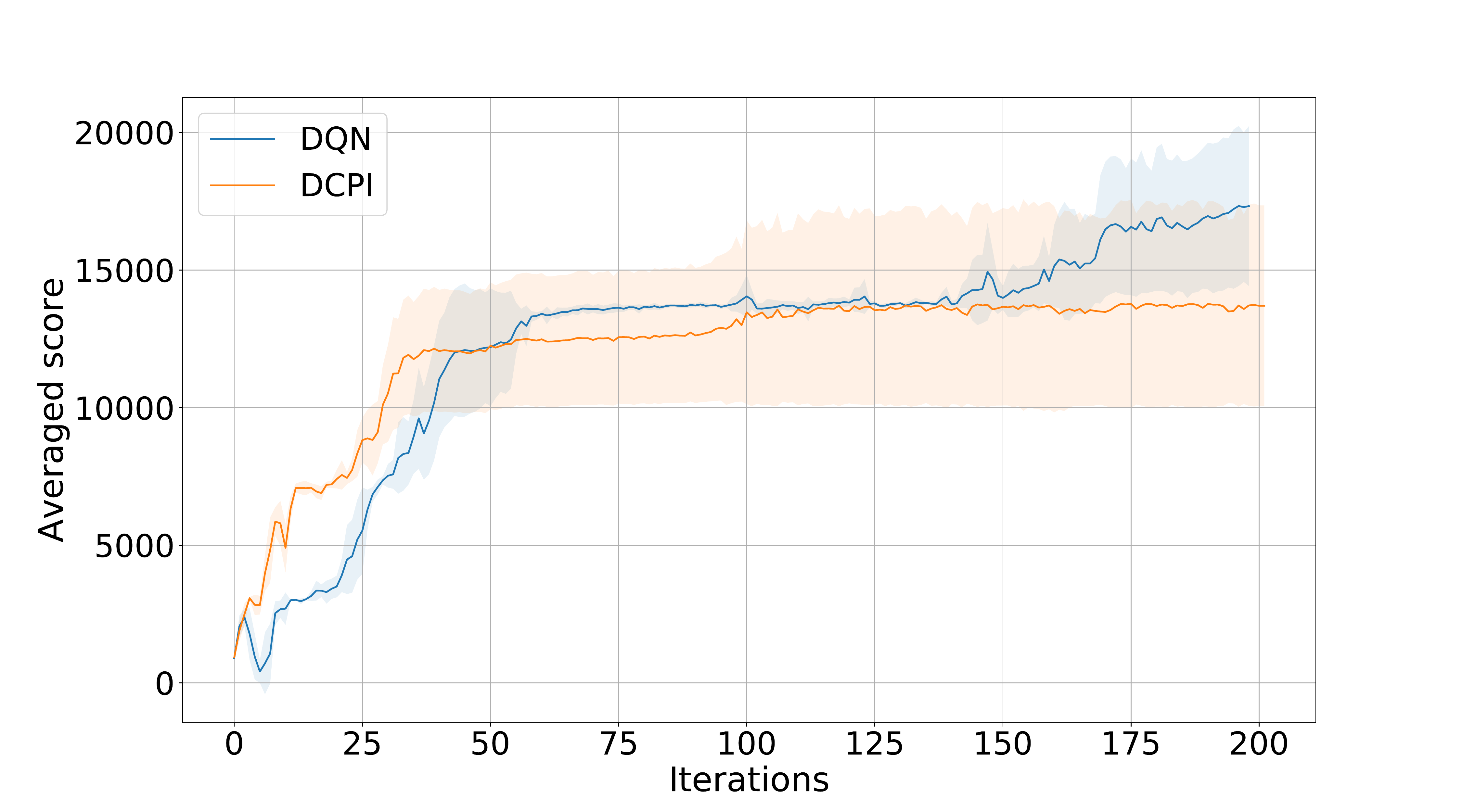} & \includegraphics[width=0.42\linewidth]{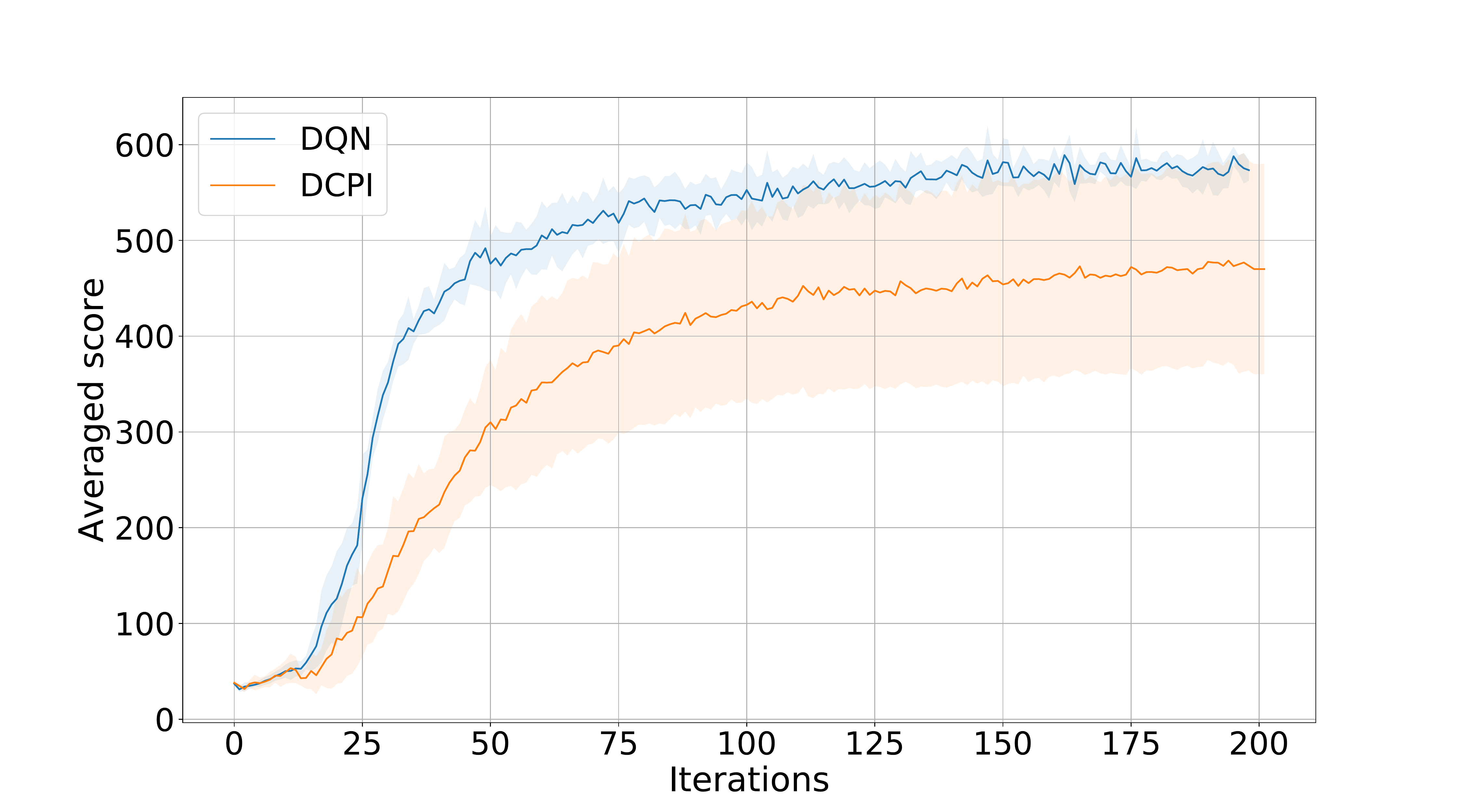}  \\
     Hero & Jamesbond \\
\end{tabular}
\caption{All averaged training scores of DCPI (orange) against DQN (blue) on the subset of Atari games (1/2).\label{fig:full}}
\end{center}
\end{figure}

\begin{figure}
\begin{center}
\begin{tabular}{cc}
     \includegraphics[width=0.42\linewidth]{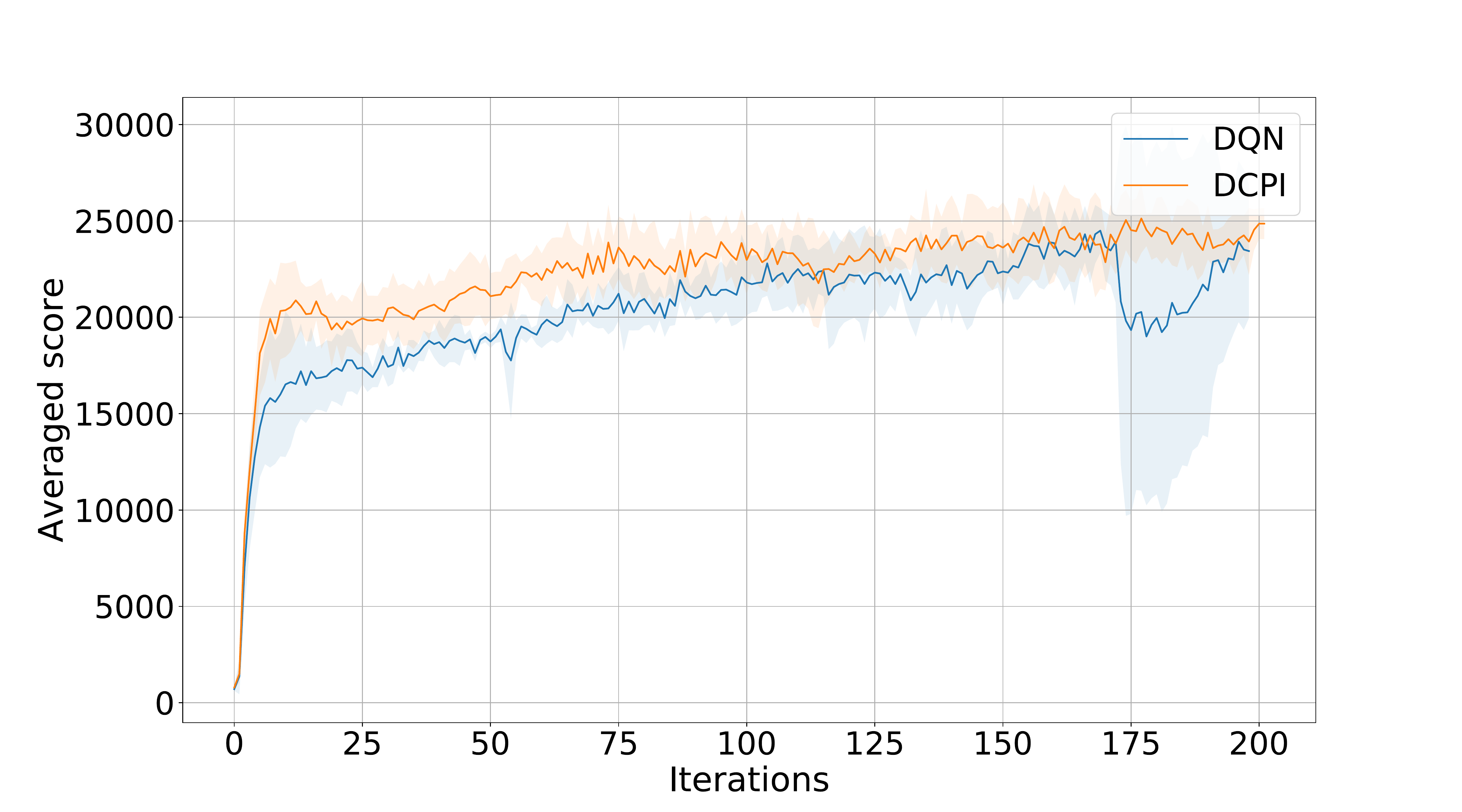} & \includegraphics[width=0.42\linewidth]{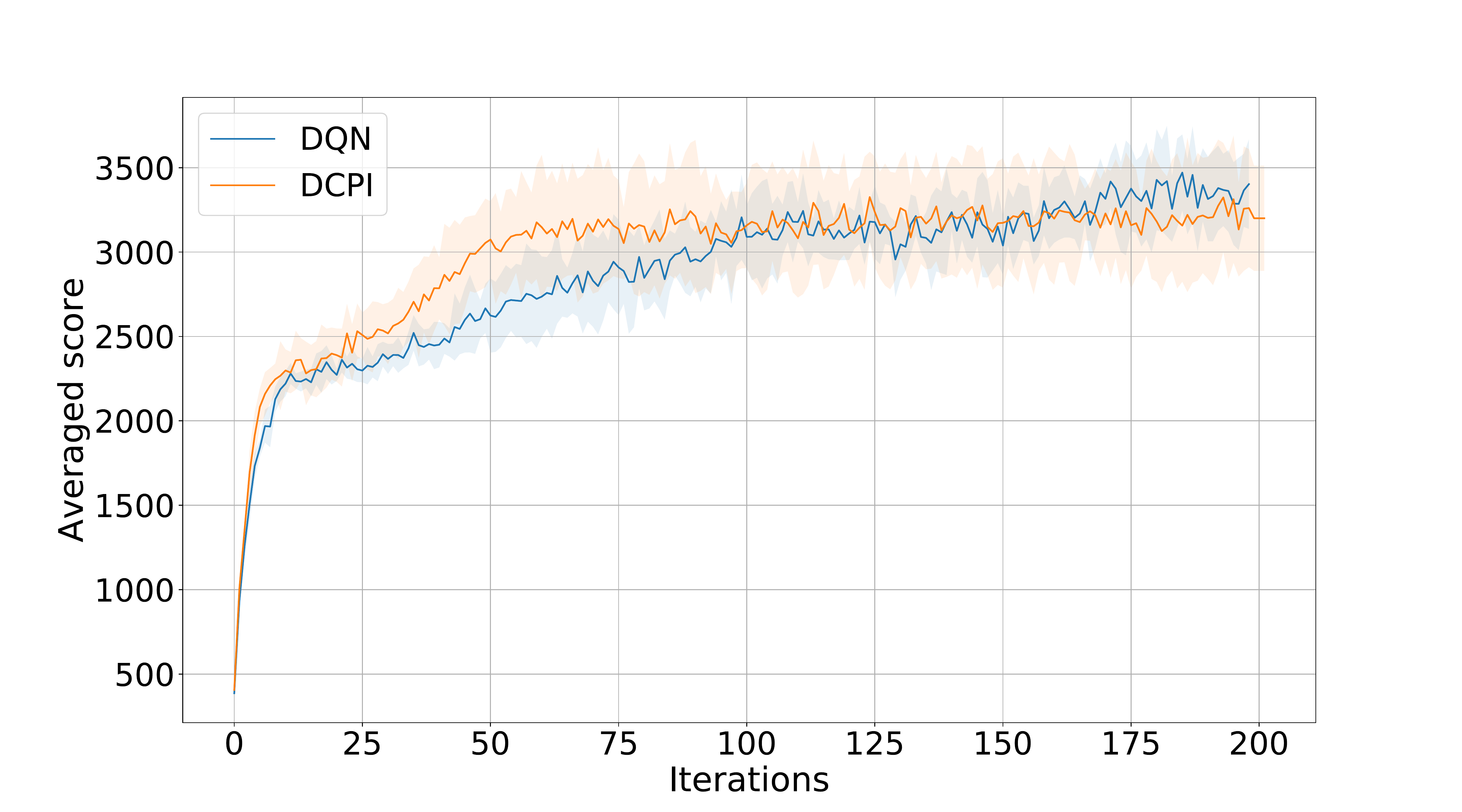}  \\
     KungFuMaster & MsPacman \\ 
     \includegraphics[width=0.42\linewidth]{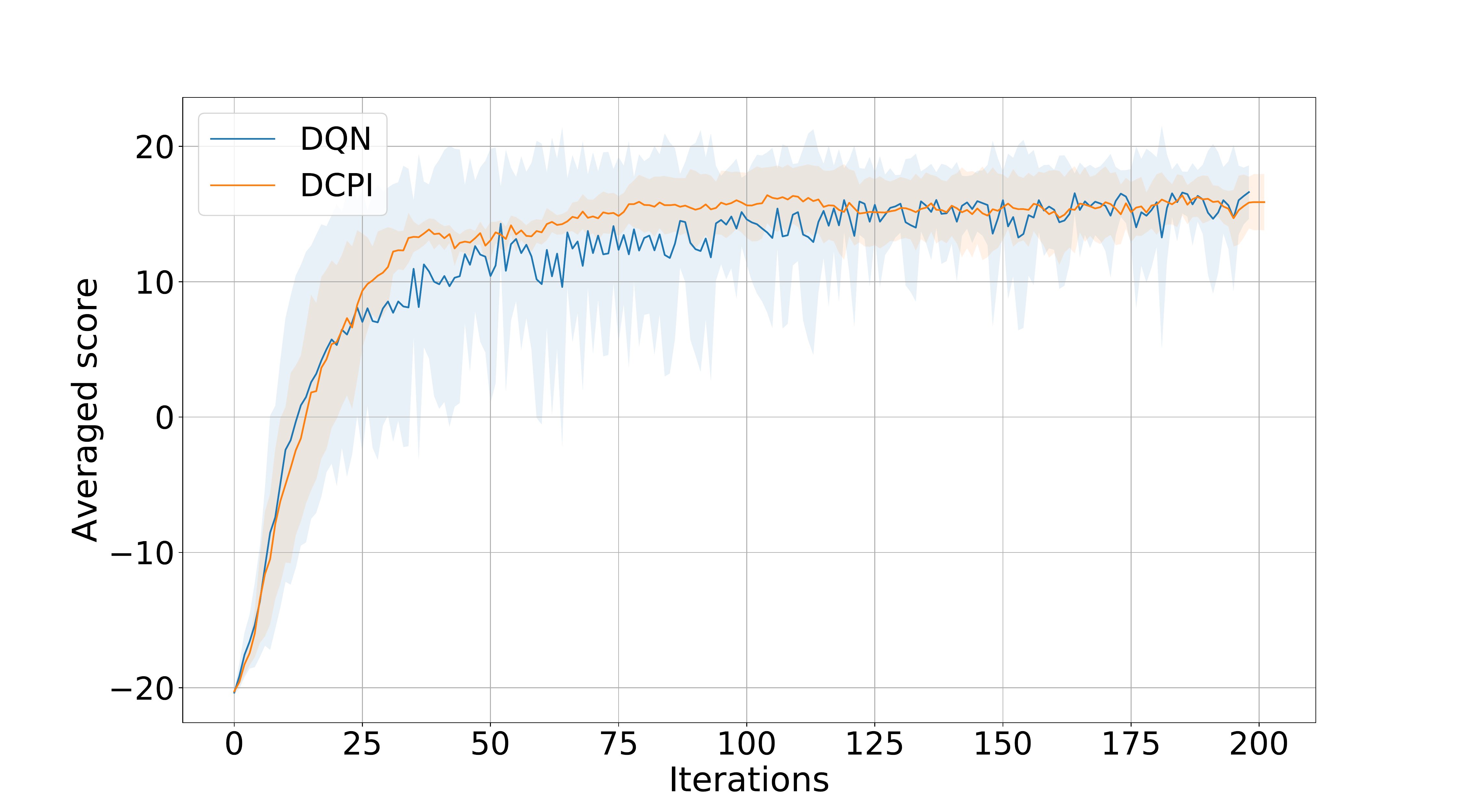} & \includegraphics[width=0.42\linewidth]{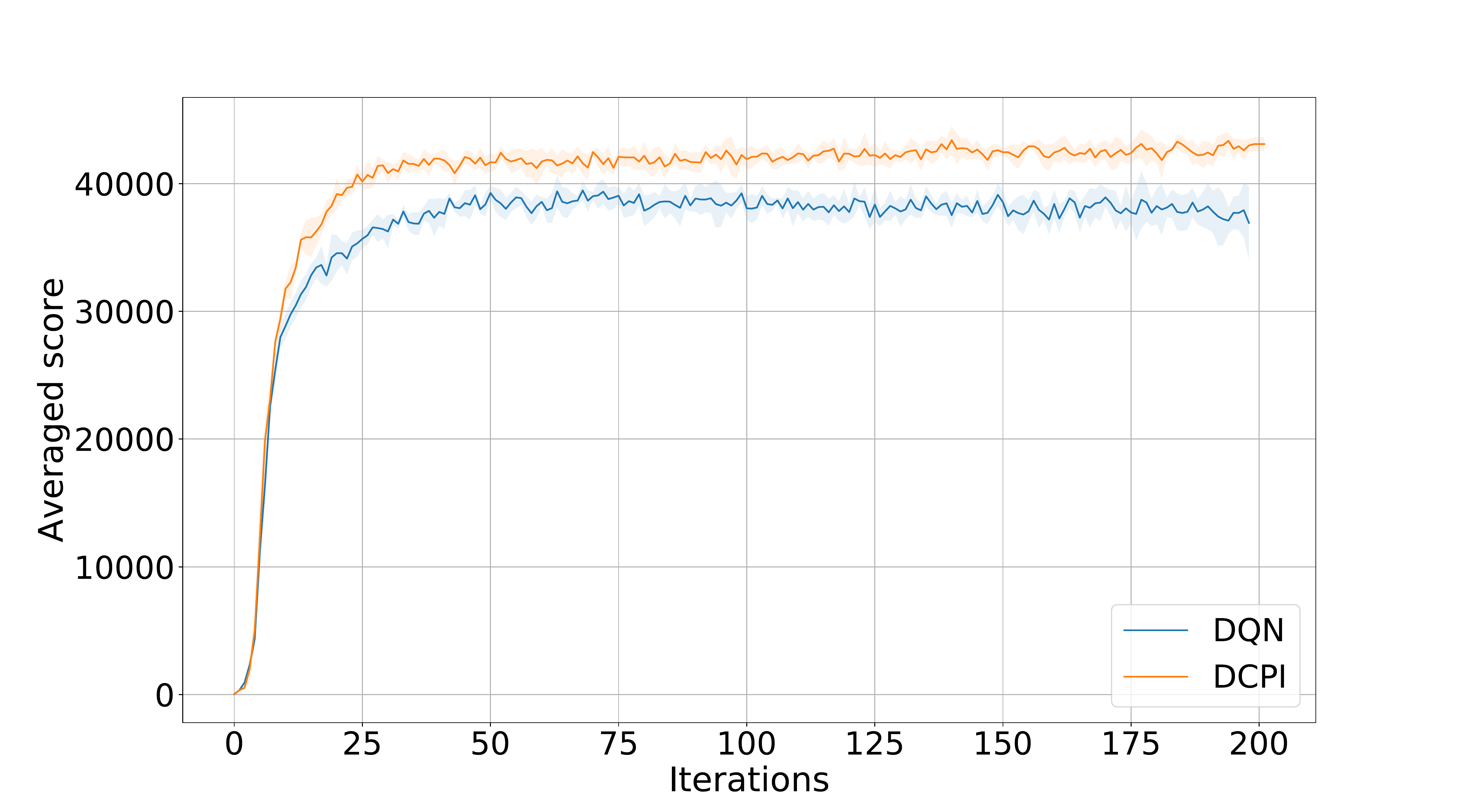}  \\
     Pong & RoadRunner \\
     \includegraphics[width=0.42\linewidth]{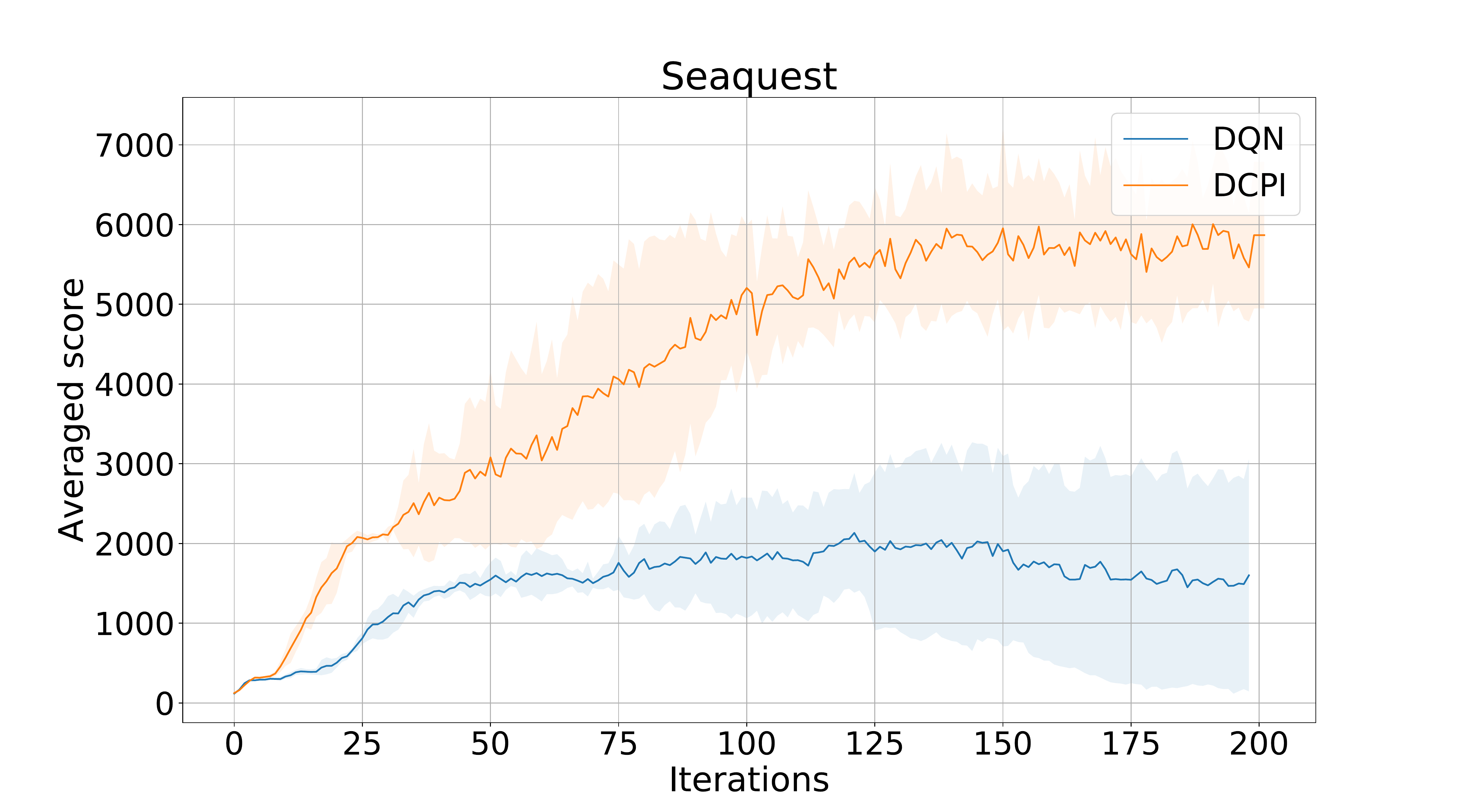} & \includegraphics[width=0.42\linewidth]{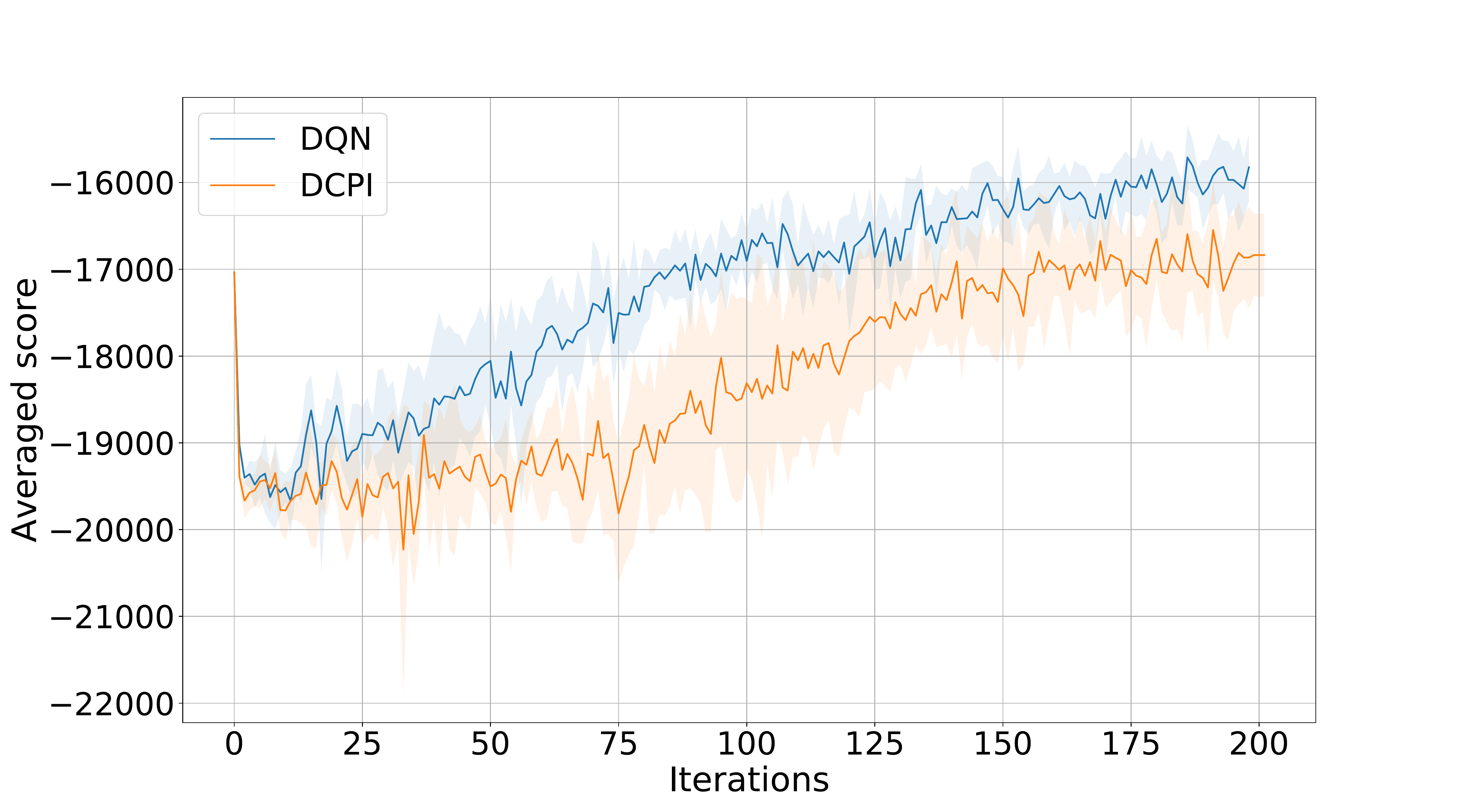}  \\
     Seaquest & Skiing \\
     \includegraphics[width=0.42\linewidth]{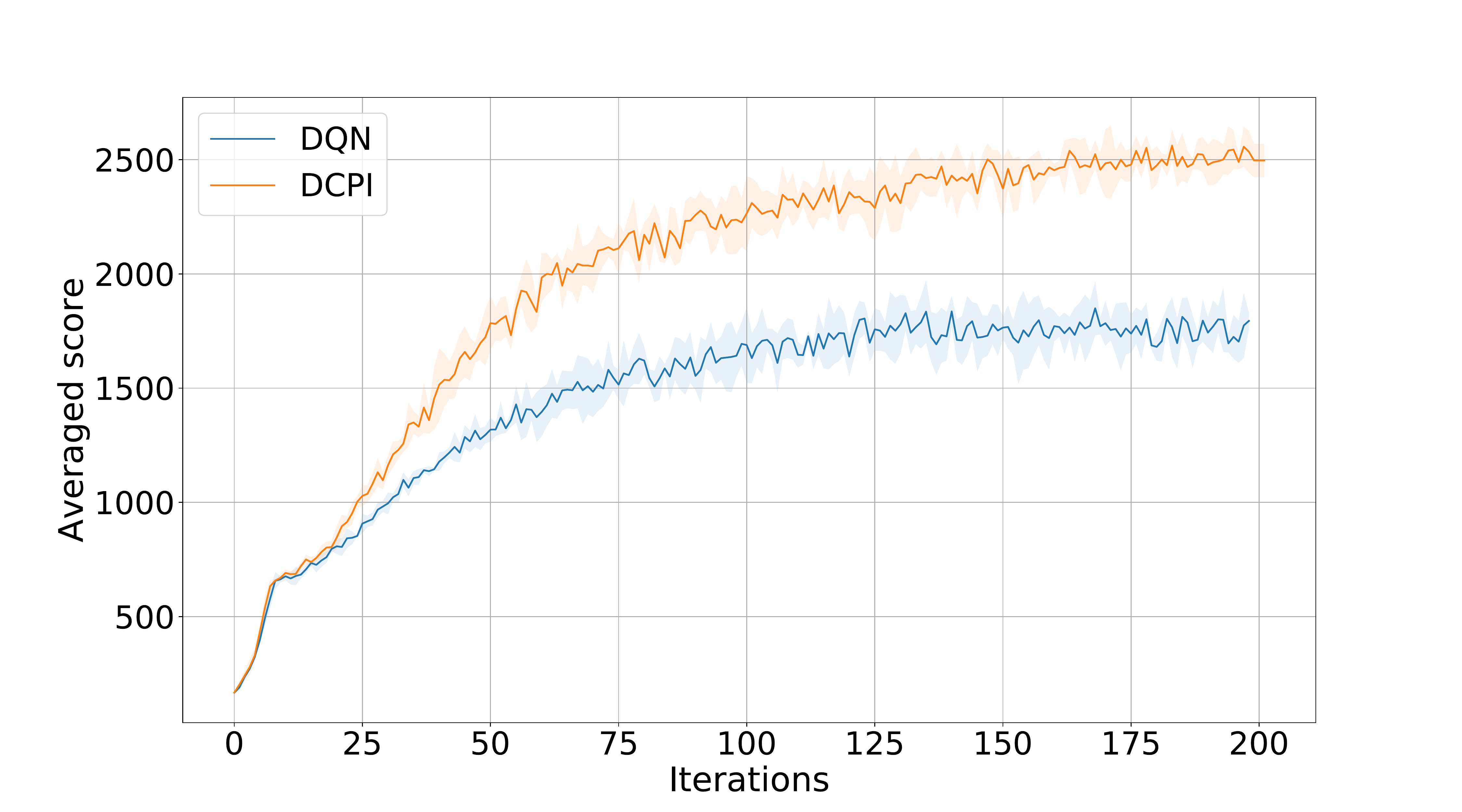} & \includegraphics[width=0.42\linewidth]{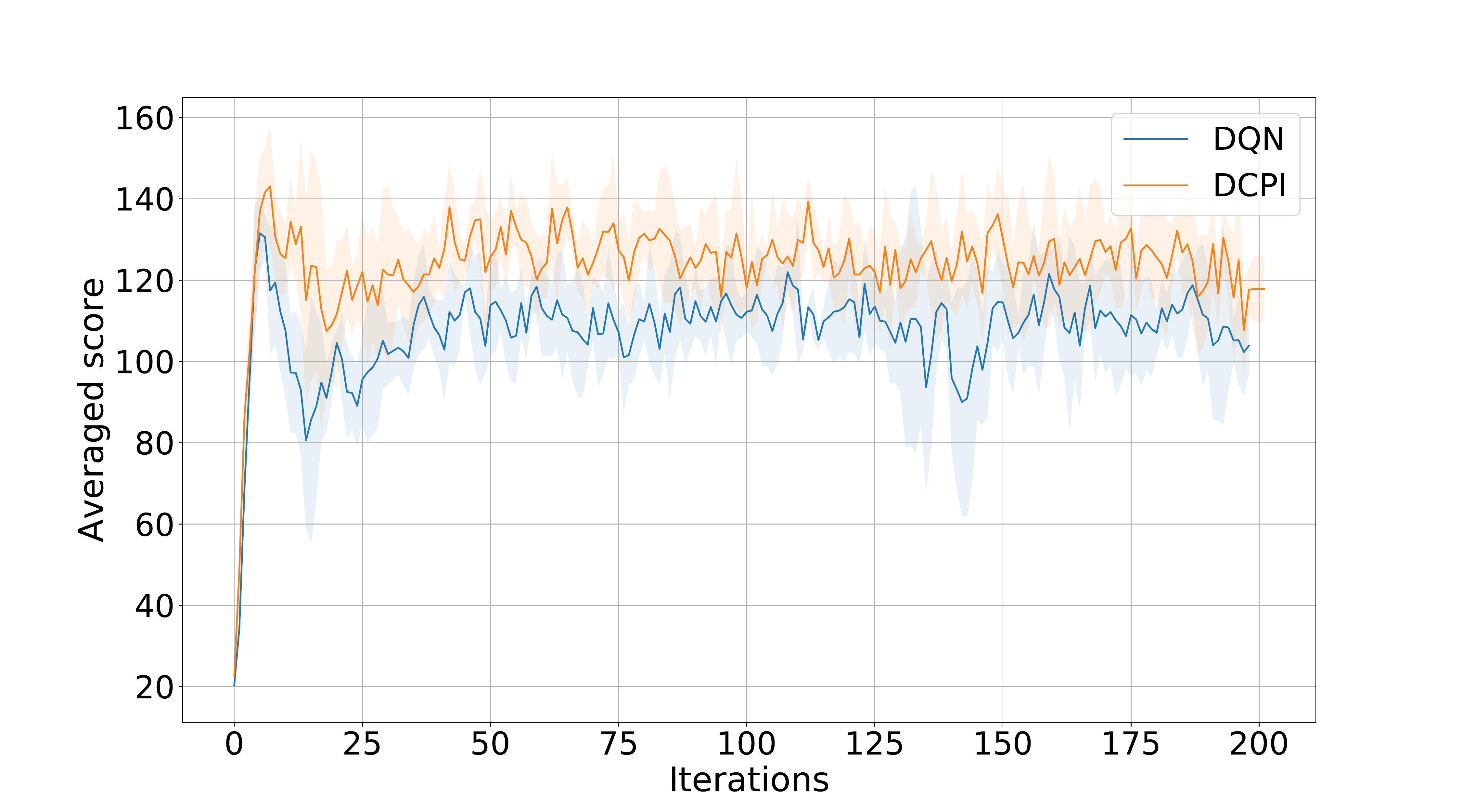}  \\
     SpaceInvaders & Tutankham \\
     \includegraphics[width=0.42\linewidth]{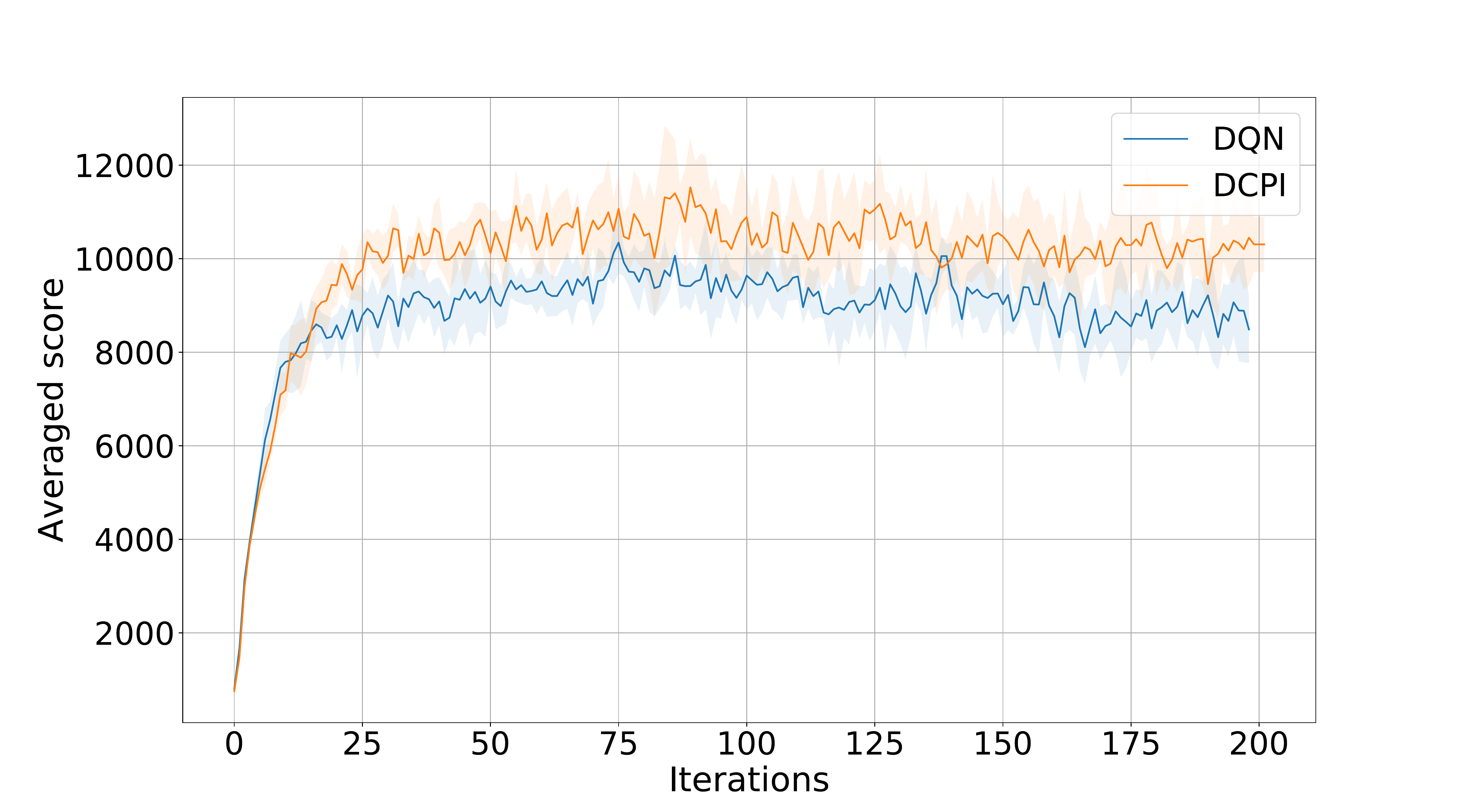} & \includegraphics[width=0.42\linewidth]{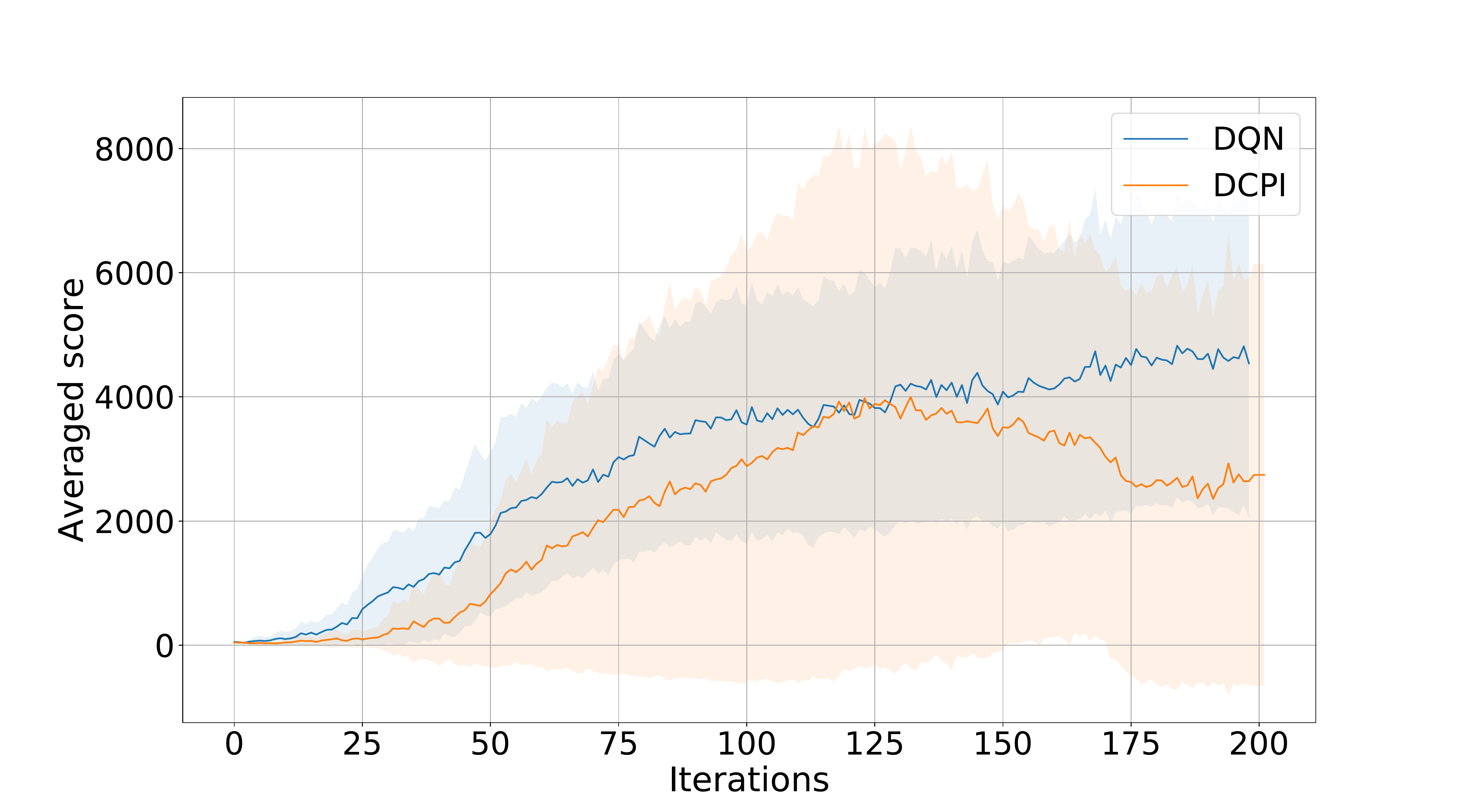}  \\
     UpNDown & Zaxxon \\
\end{tabular}
\caption{All averaged training scores of DCPI (orange) against DQN (blue) on the subset of Atari games (2/2).\label{fig:full2}}
\end{center}
\end{figure}

\fi

\end{document}